\documentclass{article}

\usepackage{iclr2027_conference}
\iclrfinalcopy
\usepackage{iftex}
\ifPDFTeX
  \usepackage[utf8]{inputenc}
  \usepackage[T1]{fontenc}
  \usepackage{times}
\else
  \usepackage{fontspec}
\fi

\usepackage{hyperref}
\usepackage{url}
\usepackage{booktabs}
\usepackage{amsfonts}
\usepackage{amsmath}
\usepackage{amssymb}
\usepackage{nicefrac}
\usepackage{microtype}
\usepackage{xcolor}
\usepackage{graphicx}
\usepackage{float}
\usepackage{enumitem}
\usepackage{amsthm}
\usepackage{placeins}
\usepackage{comment}
\usepackage{subcaption}

\graphicspath{{figures/}}

\newcommand{\R}{\mathbb{R}}

\newcommand{\NBshortk}{Nine-by-30k}
\newcommand{\NBlongk}{Nine-by-300k}
\newcommand{\RBench}{RouterBench}
\newcommand{\RTwoBench}{R2-Bench}
\newcommand{\EmbedLLM}{EmbedLLM}
\newcommand{\CarrotB}{CARROT/SPROUT}

\title{The Routing Plateau: Understanding the Accuracy Limits of LLM Routers}

\author{%
  Yifan Lu\textsuperscript{1}, Qiyue Zhang\textsuperscript{1}, Shenrun Zhang\textsuperscript{1}, Zhibo Yu\textsuperscript{1}, \\
  \textbf{Zhuang Wang\textsuperscript{2}, Hanjie Chen\textsuperscript{1}, Jiarong Xing\textsuperscript{1}} \\
  \textsuperscript{1}Rice University \quad \textsuperscript{2}Amazon \\
  \texttt{\{yl231, jxing\}@rice.edu}
}

\date{}
\hypersetup{
  hidelinks,
  pdftitle={The Routing Plateau: Understanding the Accuracy Limits of LLM Routers},
  pdfauthor={Yifan Lu, Qiyue Zhang, Shenrun Zhang, Zhibo Yu, Zhuang Wang, Hanjie Chen, Jiarong Xing}
}

\begin{document}
\maketitle
\fancyhead{}
\renewcommand{\headrulewidth}{0pt}

\begin{abstract}
LLM routing has become a popular approach to improve the cost--quality trade-off of LLM services by adaptively selecting a model for each query.
Recent work has explored a broad range of routing methods, including clustering-based routers, learned classifiers, pairwise ranking, and confidence-based approaches. 
Our extensive study of 21 routing methods across five benchmarks reveals a consistent phenomenon that we call the \textbf{routing plateau} (Fig.~\ref{fig:teaser}): 
\emph{many methods, including kNN, achieve very similar accuracy and converge to a narrow performance range that remains far below the oracle router.}
Our analysis supports a \emph{correctness-prediction bottleneck} hypothesis: current routers primarily learn global-average model performance trends rather than fine-grained, query-specific routing signals.
As a result, they collectively fail on queries that require instance-specific routing decisions.
Moreover, to understand whether the plateau can be alleviated with a better training setup, we construct a 300K-query benchmark (\NBlongk{}).
More data, larger encoders, and end-to-end fine-tuning improve eight routers by $1.24$~pp on average, but leave the plateau largely intact.
These findings suggest that further progress may require inputs beyond the query itself, such as partial output trajectories that reveal how models attempt the task.
\end{abstract}

\begin{figure}[h]
\centering
\includegraphics[width=\linewidth]{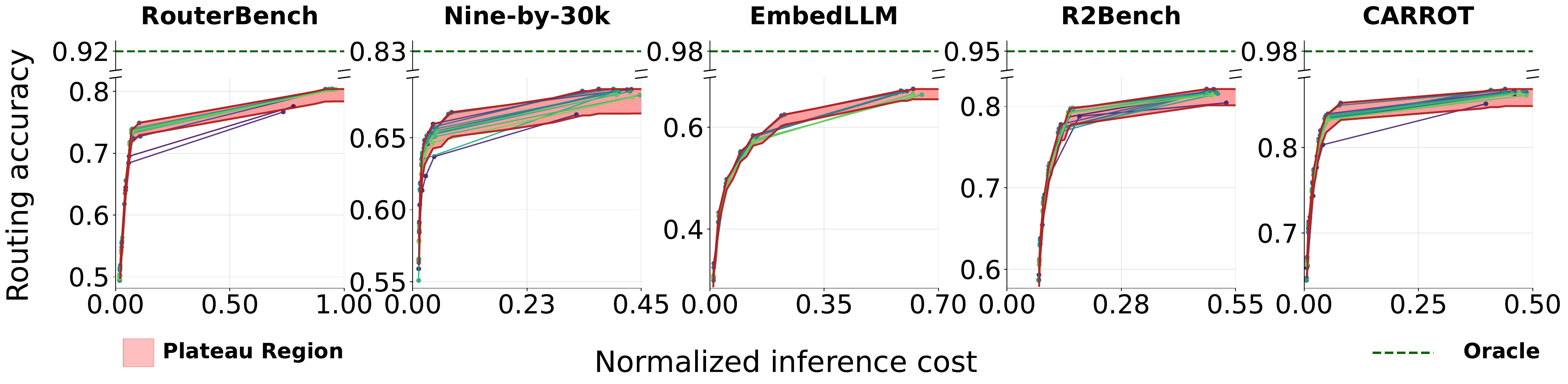}
\caption{Routing accuracy vs.\ normalized inference cost. All routers from heterogeneous method families collapse onto a narrow \textit{plateau} (red band) well below the per-query oracle (green dashed).}
\label{fig:teaser}
\end{figure}

\section{Introduction}
\label{sec:introduction}

Query routing across models has become a common approach for improving the cost--quality trade-off of LLM services.
An LLM router dynamically selects an appropriate model for each query, routing simpler queries to smaller, cheaper, or more specialized models while reserving expensive frontier models for harder queries~\citep{hu2024router_bench,somerstep2025carrot,ong2025routellm,xue2026r2_router}.
Recent work has proposed a wide range of routing methods, including similarity search over training examples~\citep{lu2023zooter_lite,somerstep2025carrot}, learned classifiers or regressors for model correctness prediction~\citep{ong2025routellm,Sakota2024forc,liu2024optllm,notdiamond2025rorf}, pairwise model ranking~\citep{frick2025p2l,zhao2024elo_router}, and confidence- or lookahead-based methods~\citep{ding2024hybrid_llm,woisetschlager2025mess_plus,zhang2025avengers_pro}. 
Each new routing method typically reports advantages over prior routers, suggesting steady progress in router design. 
However, because these studies often differ in model pools, benchmarks, training data, and evaluation protocols~\citep{lu2025routerarena}, it remains unclear whether these reported gains reflect fundamental improvements in routing capability or artifacts of evaluation setup.

In this paper, we evaluate 21 routing methods across five representative routing benchmarks.
Our study reveals a phenomenon we call the \textbf{routing plateau}: \emph{despite their diverse designs, many routers, including kNN-style routers, converge to the same accuracy ceiling, which remains far below that of the oracle router}~\citep{lai2026equirouter}.
We unpack this plateau through three more specific empirical observations.
\textbf{(1) Similar top-end accuracy:} Although a few strong routers outperform other methods, their accuracies are highly similar and converge to a narrow performance band. 
For example, on \RBench{}, the top 15 routers differ by only 0.23 percentage points (pp) in accuracy.
\textbf{(2) Strong kNN-style routers:} simple similarity-based approaches such as kNN remain consistently competitive with more sophisticated routers across benchmarks.
For instance, kNN ranks among the top-2 methods on all five benchmarks.
\textbf{(3) A persistent oracle gap:} all existing routers remain far below the oracle router, showing that the key challenge is identifying the correct model from the query.
For example, the best router trails the Oracle by 10--30 pp across the five benchmarks.

Our further analysis supports a hypothesis of the \textbf{correctness-prediction bottleneck}:
Routers must predict which models will answer a query correctly, but often struggle to learn this signal from available data and query representations.
Specifically, current routers tend to learn coarse, global patterns of model capability rather than fine-grained, instance-level differences in correctness.
For many queries, however, model correctness does not always align with overall capability: models with higher average performance are not consistently the most reliable on individual instances.
As a result, although routers may differ in routing decisions, their shared reliance on global-model-capability signals causes them to converge to a narrow accuracy ceiling far below that of the oracle router.
This also explains why kNN-style routers remain competitive: simple similarity-based methods can already capture much of this coarse routing signal.

Finally, we ask whether the plateau reflects limitations of the training regime.
To test that, we construct \NBlongk{}, expanding the training set by 10 times to 302{,}371 labeled queries.
We examine data scaling, larger query encoders, and end-to-end fine-tuning~\citep{chen2024router_dc,feng2025graph_router} across eight router designs.
Combining these interventions improves accuracy by $1.24$~pp on average and up to $2.13$~pp, with most of the gain coming from more data and fine-tuning.
Yet routers remain closely clustered and far below the oracle.
These interventions improve how routers learn from queries, but still require them to predict correctness before observing any model response.
These findings suggest that further progress may require richer routing inputs, such as partial output trajectories that reveal how models attempt the query.

In summary, this paper makes the following contributions:
\vspace{-1mm}
\begin{itemize}[leftmargin=1.5em, topsep=0.2em, itemsep=0.2em, parsep=0pt]
    \item We evaluate 21 LLM routing methods across five benchmarks under a unified experimental setup and reveal the routing plateau phenomenon.

    \item We provide evidence for a correctness-prediction bottleneck behind the plateau: current routers tend to learn coarse, global patterns of model capability rather than fine-grained routing signals.

    \item We construct \NBlongk{}, a routing benchmark with 300k training queries and 2.8 million query--model correctness labels, enabling us to test the role of data scaling, encoder capacity, and end-to-end fine-tuning in the plateau.

    \item We show that these training interventions yield modest gains while leaving the plateau largely intact, motivating routing inputs that provide evidence of model behavior beyond the query alone.
\end{itemize}

\section{Related Work}
\label{sec:related_work}

\paragraph{LLM routing methods.}
Recent routers fall into a few design families: \emph{classifier-style} predictors of per-model correctness \citep{ong2025routellm,Sakota2024forc,liu2024optllm,notdiamond2025rorf,jin2025radialrouter}, \emph{retrieval} routers that copy the best neighbor from training \citep{lu2023zooter_lite,somerstep2025carrot}, \emph{pairwise-ranking} routers \citep{frick2025p2l,zhao2024elo_router}, \emph{latent-factor} and \emph{IRT} models \citep{zhuang2024embedllm,song2025irt_router}, and \emph{contrastive} routers that pull a query toward the model answering it correctly \citep{chen2024router_dc,lai2026equirouter}. 
Cascades, orchestration, bandit-style, multi-round, and online-serving routers \citep{ding2024hybrid_llm,zhang2023model_spider,zhang2025avengers_pro,woisetschlager2025mess_plus,Li2010linucb,zhang2025router_r1,wu2025online_routing} sit at the edge of the single-shot setting and serve as reference points. 
Each new method is usually evaluated against a small set of baselines on a single benchmark under its own encoder and experimental setup. 

\paragraph{Routing benchmarks and analysis.}
Several public benchmarks are available: RouterBench~\citep{hu2024router_bench}, EmbedLLM~\citep{zhuang2024embedllm}, R2-Bench~\citep{xue2026r2_router}, CARROT/SPROUT~\citep{somerstep2025carrot}, RouterArena~\citep{lu2025routerarena}, and LLMRouterBench~\citep{li2026llmrouterbench}. 
On the analysis side, \citet{lai2026equirouter} show that trained routers tend to collapse onto a small set of preferred models; 
\citet{li2026llmrouterbench} report that many recent and even commercial routers fail to reliably outperform a simple baseline; 
and \citet{li2025knn_router} argue that a well-tuned frozen-embedding kNN already matches state-of-the-art learned routers across diverse tasks. 
Our work offers a deeper analysis of the routing plateau, understanding its reasons and identifying potential paths forward.

\section{Preliminaries and Problem Setup}
\label{sec:prelim}

\paragraph{LLM Routing.}
A routing instance is a pair $(\mathcal{Q}, \mathcal{P})$, where $\mathcal{Q}$ is a query distribution and $\mathcal{P} = \{1, \ldots, K\}$ is a fixed pool of $K$ candidate LLMs. 
Each $(q_i, m) \in \mathcal{Q} \times \mathcal{P}$ is associated with a binary correctness label $Y_{i,m} \in \{0,1\}$ and a real-valued per-call cost $c_{i,m} \ge 0$. 
A router is a function $r$ that maps a query to a model, $\hat m_i = r(q_i) \in \mathcal{P}$. 
Most routers decompose into a \emph{predictor} that produces scores $\hat p_{i,m} \in \R$ approximating $\Pr(Y_{i,m}=1 \mid q_i)$ and a \emph{selector} that returns $\hat m_i = \arg\max_m \hat p_{i,m}$.

\paragraph{Cost-aware utility.}
For $\lambda \in [0,1]$, the cost-utility scalarization of choosing model $m$ on query $q_i$ is $
U(\lambda;\, q_i, m) \;=\; (1-\lambda)\,\hat p_{i,m} \;-\; \lambda\,\tilde c_{i,m},
$
generalizing the metric used in \RBench{} and \RTwoBench{}~\citep{hu2024router_bench, xue2026r2_router}. 
$\lambda = 0$ recovers pure accuracy maximization; $\lambda = 1$ is pure cost minimization. Throughout the paper, evaluation cells are indexed by the triple  $(\mathcal{Q}, \mathcal{P}, \lambda)$.

\paragraph{Routing accuracy and the oracle.}
The accuracy of router $r$ on a test set $\mathcal{Q}_{\mathrm{test}}$ of size $N$ is
$A_r \;=\; \frac{1}{N} \sum_{i=1}^{N} Y_{i,\,r(q_i)}.
$
If a router could see the correctness labels of every model before choosing, it would pick a correct one, giving
$
A_{\mathrm{oracle}}
\;=\;
\frac{1}{N} \sum_{i=1}^{N} \max_{m \in \mathcal{P}} Y_{i,m}.
$
$A_{\mathrm{oracle}}$ depends only on $(\mathcal{Q}, \mathcal{P})$ and not on which router we use. 
It is computed from labels that a real router never sees at decision time, so it sits as an upper bound rather than something deployed systems can match.

\paragraph{Concordance.}
It assesses whether the router generally ranks correct models above incorrect ones. For each test query $q_i$, let $\mathcal{S}_i = \{(m,n) \in \mathcal{P}^2 : Y_{i,m}=1,\;Y_{i,n}=0\}$ contain all pairs of models where $m$ answers that query correctly and $n$ does not. We compute
{\begin{equation}
\operatorname{Conc}(r) =
\frac{\displaystyle\sum_{i=1}^{N}\sum_{(m,n)\in\mathcal{S}_i}
\left[\mathbf{1}\{\hat p_{i,m}>\hat p_{i,n}\}
+\tfrac{1}{2}\mathbf{1}\{\hat p_{i,m}=\hat p_{i,n}\}\right]}
{\displaystyle\sum_{i=1}^{N}|\mathcal{S}_i|}.
\label{eq:concordance}
\end{equation}
Thus, concordance is the fraction of correct--incorrect model pairs ranked correctly.

\section{The Routing Plateau Across Routers and Benchmarks}
\label{sec:experiments}

In this section, we characterize the routing plateau through a large-scale empirical study covering 21 routing methods across five representative benchmarks.

\subsection{Routers, Benchmarks, and Evaluation Methods}
\label{sec:exp:benchmarks}

\paragraph{Routing methods and encoders.}
We evaluate 21 routers in total, including 18 methods that span the major design families proposed from 2023 to 2026:
classifier-style routers \citep{ong2025routellm,Sakota2024forc,liu2024optllm,notdiamond2025rorf},
retrieval routers \citep{lu2023zooter_lite,somerstep2025carrot},
ranking and pairwise routers \citep{frick2025p2l,zhao2024elo_router},
latent-factor and IRT routers \citep{song2025irt_router,zhuang2024embedllm},
contrastive routers \citep{chen2024router_dc,lai2026equirouter},
expert-orchestration and cascade routers \citep{zhang2025avengers_pro,zhang2023model_spider,ding2024hybrid_llm,woisetschlager2025mess_plus},
and bandit/cost-aware routers \citep{Li2010linucb}.
We additionally include three classical baselines: \texttt{mlp\_cn} and \texttt{knn} \citep{hu2024router_bench}, and \texttt{kmeans} \citep{jitkrittum2025universalmodelroutingefficient}.

We use six query encoders across three tiers:
sentence-transformer encoders \textsc{MiniLM-L6}~\citep{wang2020minilm} and \textsc{MPNet-base}~\citep{song2020mpnet};
BERT-family encoders \textsc{BGE-base}~\citep{xiao2023bge}, \textsc{ModernBERT-base} and \textsc{ModernBERT-large}~\citep{warner2024modernbert};
{and a decoder-only LLM as encoder \textsc{Qwen2.5-0.5B}~\citep{qwenteam2024qwen25}.}
The six encoders span $384$ to $1024$ output dimensions and three orders of magnitude of parameter count (App.~\ref{app:experiments}).

\paragraph{Benchmarks.}
We evaluate routers on five benchmarks. Four of them are public LLM-routing benchmarks: \RBench{}~\citep{hu2024router_bench}, \RTwoBench{}~\citep{xue2026r2_router}, \EmbedLLM{}~\citep{zhuang2024embedllm}, and \CarrotB{}~\citep{somerstep2025carrot}.
They are large enough to train a frozen-embedding prediction head, but are too small for reliable data scaling analysis or end-to-end encoder fine-tuning.
To close this gap, we construct a new benchmark with two paired subsets.
{\NBshortk{} provides 23{,}027 training queries labeled for nine models for the main comparison.}
{\NBlongk{} expands this to 302{,}371 training queries for data scaling and encoder fine-tuning.}
We defer the construction details of our benchmark to App.~\ref{app:dataset}.

\paragraph{Evaluation methods.}
{Each router configuration is a (encoder, method) pair, where the encoder produces query embeddings and the method defines the routing rule on embeddings.}
We use the same train/validation/test split for all router configuration.
For each (benchmark, encoder, method) setting, we sweep architectures and hyperparameters, selecting the variant with the highest validation accuracy at $\lambda{=}0$.

\subsection{The Routing Plateau Phenomenon}
\label{sec:exp:results}

\begin{figure}
    \centering
    \includegraphics[width=\linewidth]{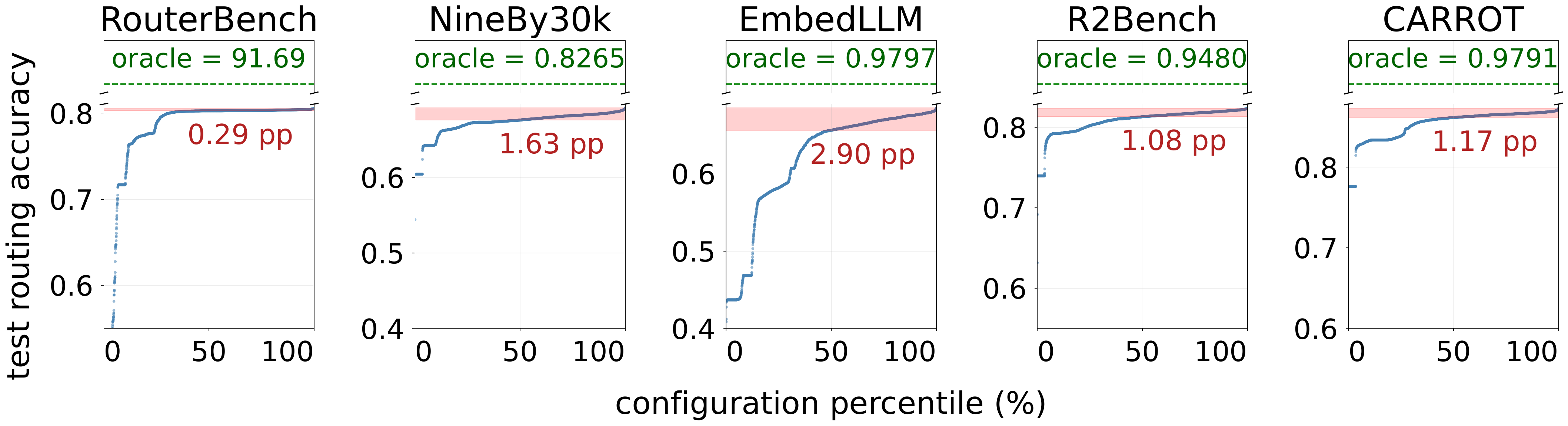}
    \caption{Test routing accuracy at $\lambda{=}0$ across all sorted configurations. Red bands mark the accuracy ranges of the highest-scoring half of configurations. Green dashed lines mark oracle accuracy.}
    \label{fig:ceiling_per_run}
    \vspace{-3mm}
\end{figure}

\paragraph{Extensive configuration sweep: the plateau is broad.}
{Our sweep covers a broad grid of router configurations.}
{Fig.~\ref{fig:ceiling_per_run} sorts all configurations by their test accuracy within each benchmark.}
{The resulting distribution shows a broad plateau: on all five benchmarks, at least half of the configurations fall within a short range of the highest observed test accuracy.}
{Thus, the plateau is reached by many configurations, spanning different methods, and encoders landing in the same band.}
{However, even the best configurations remain $10.6$--$29.4$~pp below the oracle router: they still select models whose responses score lower than the best available response.}

\paragraph{{Top-tier methods are nearly indistinguishable.}}
We next ask whether any routing method consistently separates itself from the rest.
Tab.~\ref{tab:main:lambda0_all} reports test routing accuracy at $\lambda{=}0$ on all benchmarks, with results averaged across encoders.
It shows that the leading methods form a tight cluster: the top-5 methods differ by only $0.22$~pp on average.
{Paired bootstrap tests among the top-five methods found no significant accuracy differences in the comparisons (App.~\ref{app:significance}).}
{Meanwhile, half of all methods lie within $0.08$--$1.13$~pp of the best, under $5\%$ of the oracle gap between the best router and the oracle.}
{These results suggest that the leading methods have reached a narrow accuracy plateau, with none substantially outperforming the others or approaching oracle accuracy.}

\begin{table}[t]
  \centering
  \footnotesize
  \setlength{\tabcolsep}{2pt}
  \caption{{Test routing accuracy (\%), averaged over encoders. Bold marks the best result per column.}}
  \label{tab:main:lambda0_all}
  \resizebox{\linewidth}{!}{%
  \begin{tabular}{lcccccc}
    \toprule
     & \RBench{} & \NBshortk{} & \EmbedLLM{} & \RTwoBench{} & \CarrotB{} & Avg \\
    \midrule
    Zooter~\citep{lu2023zooter_lite}              & {$80.38$} & $\mathbf{68.38}$ & $\mathbf{67.51}$ & $\mathbf{82.12}$ & {$86.64$} & $\mathbf{77.01}$ \\
    kNN~\citep{hu2024router_bench}                                            & $\mathbf{80.39}$ & {$68.36$} & {$67.16$} & {$82.08$} & {$86.78$} & $76.95$ \\
    Elo~\citep{zhao2024elo_router}                 & {$80.36$} & {$68.34$} & {$67.02$} & {$82.08$} & $\mathbf{86.85}$ & $76.93$ \\
    {Avengers-Pro~\citep{zhang2025avengers_pro}}     & {$80.34$} & {$68.31$} & {$66.66$} & {$81.89$} & {$86.75$} & {$76.79$} \\
    {EquiRouter~\citep{lai2026equirouter}}           & {$80.36$} & {$68.24$} & {$67.08$} & {$81.80$} & {$86.46$} & {$76.79$} \\
    KMeans~\citep{jitkrittum2025universalmodelroutingefficient}                                         & {$80.35$} & $\mathbf{68.38}$ & $66.67$ & {$81.87$} & {$86.64$} & $76.78$ \\
    {MESS+~\citep{woisetschlager2025mess_plus}}      & {$80.35$} & {$68.20$} & {$66.99$} & {$81.69$} & {$86.61$} & {$76.77$} \\
    CARROT~\citep{somerstep2025carrot}             & {$80.30$} & {$68.22$} & {$66.98$} & $81.54$ & $86.37$ & $76.68$ \\
    CP-Router~\citep{su2025cprouteruncertaintyawarerouterllm}                                      & {$80.29$} & {$68.22$} & $66.09$ & {$81.93$} & {$86.65$} & $76.64$ \\
    P2L~\citep{frick2025p2l}                       & {$80.31$} & $67.95$ & $66.38$ & $81.56$ & $86.32$ & $76.50$ \\
    {MLP~\citep{hu2024router_bench}}                                            & {$80.29$} & {$67.98$} & {$66.51$} & {$81.38$} & {$86.12$} & {$76.46$} \\
    {EmbedLLM~\citep{zhuang2024embedllm}}            & {$80.32$} & {$67.81$} & {$66.05$} & {$81.40$} & {$86.34$} & {$76.38$} \\
    {ModelSpider~\citep{zhang2023model_spider}}      & {$80.31$} & {$67.61$} & {$65.58$} & {$81.58$} & {$86.40$} & {$76.29$} \\
    RoRF~\citep{notdiamond2025rorf}                & $77.57$ & $67.86$ & $64.43$ & $81.60$ & $86.26$ & $75.54$ \\
    TagRouter~\citep{Sakota2024forc}               & {$80.34$} & {$67.97$} & $61.08$ & $81.41$ & $84.82$ & $75.12$ \\
    LinUCB~\citep{Li2010linucb}                    & $73.41$ & $67.76$ & $65.76$ & $80.72$ & $85.87$ & $74.70$ \\
    {RouterDC~\citep{chen2024router_dc}}             & {$80.17$} & {$67.54$} & {$58.16$} & {$80.22$} & {$83.13$} & {$73.84$} \\
    {MIRT~\citep{song2025irt_router}}                                           & {$76.36$} & {$66.64$} & {$59.20$} & {$80.94$} & {$84.88$} & {$73.60$} \\
    {IRT-Router~\citep{song2025irt_router}}          & {$77.49$} & {$66.49$} & {$52.06$} & {$81.04$} & {$84.78$} & {$72.37$} \\
    HybridLLM~\citep{ding2024hybrid_llm}           & $71.67$ & $60.46$ & $46.87$ & $73.99$ & $77.62$ & $66.12$ \\
    OptLLM~\citep{liu2024optllm}                   & $76.68$ & $66.61$ & --- & $80.43$ & $85.12$ & --- \\
    \midrule
    Top-5 router spread (max$-$min) & 0.04 & 0.07 & 0.52 & 0.23 & 0.21 & 0.22 \\
    \bottomrule
  \end{tabular}%
  }
\vspace{-3mm}
\end{table}

\paragraph{The plateau exists beyond pure accuracy.}
We next examine whether the plateau remains under cost-aware routing objectives.
{Tab.~\ref{tab:main:nb30k_utility} reports the realized cost-utility $U(\lambda)$ for the 16 calibrated methods on \NBshortk{} at $\lambda{>}0$.}
{The same near-tie pattern persists: the mean top-five cost-utility spread is only $0.18$ across the four cost tradeoffs.}
{This suggests that the plateau persists when balancing accuracy against inference cost, with no leading method offering a substantial cost-utility advantage over the others.}
The pattern generalizes beyond \NBshortk{} (App.~\ref{app:additional_results:per_benchmark})

\begin{table}[t]
  \centering
  \footnotesize
  \setlength{\tabcolsep}{5pt}
  \caption{{Cost-utility on \NBshortk{} at $\lambda{>}0$: $U(\lambda){=}(1{-}\lambda)\hat{s}{-}\lambda\tilde{c}$, mean over six encoders. Sorted by $\lambda{=}0.5$. All numbers are in percent (\%).}
  }
  \label{tab:main:nb30k_utility}
  \begin{tabular}{lccccc}
    \toprule
     & $\lambda{=}0.2$ & $\lambda{=}0.5$ & $\lambda{=}0.7$ & $\lambda{=}0.9$ & Avg \\
    \midrule
    Zooter~\citep{lu2023zooter_lite}              & {$\mathbf{51.97}$} & {$\mathbf{31.25}$} & {$\mathbf{17.82}$} & {$4.76$} & $\mathbf{26.45}$ \\
    EquiRouter~\citep{lai2026equirouter}           & {$51.80$} & {$31.13$} & {$17.75$} & {$4.75$} & $26.36$ \\
    MESS+~\citep{woisetschlager2025mess_plus}      & {$51.59$} & {$31.06$} & {$17.72$} & {$4.72$} & $26.27$ \\
    P2L~\citep{frick2025p2l}                       & {$51.18$} & {$31.06$} & {$17.73$} & {$4.67$} & $26.16$ \\
    KMeans~\citep{jitkrittum2025universalmodelroutingefficient}                                         & {$51.91$} & {$31.05$} & {$17.76$} & {$4.83$} & $26.39$ \\
    CARROT~\citep{somerstep2025carrot}             & {$51.50$} & {$31.02$} & {$17.42$} & {$4.59$} & $26.13$ \\
    Elo~\citep{zhao2024elo_router}                 & {$51.64$} & {$31.01$} & {$17.42$} & {$4.59$} & $26.16$ \\
    Avengers-Pro~\citep{zhang2025avengers_pro}     & {$51.56$} & {$30.99$} & {$17.39$} & {$4.57$} & $26.13$ \\
    kNN~\citep{hu2024router_bench}                                            & {$51.89$} & {$30.94$} & {$17.70$} & {$\mathbf{4.84}$} & $26.34$ \\
    MLP~\citep{hu2024router_bench}                                            & {$51.39$} & {$30.93$} & {$17.70$} & {$4.74$} & $26.19$ \\
    RoRF~\citep{notdiamond2025rorf}                & {$51.08$} & {$30.88$} & {$17.14$} & {$4.53$} & $25.91$ \\
    EmbedLLM~\citep{zhuang2024embedllm}            & {$50.93$} & {$30.78$} & {$17.56$} & {$4.66$} & $25.98$ \\
    OptLLM~\citep{liu2024optllm}                   & {$50.10$} & {$29.89$} & {$17.10$} & {$4.68$} & $25.44$ \\
    CP-Router~\citep{su2025cprouteruncertaintyawarerouterllm}                                      & {$50.41$} & {$28.67$} & {$16.14$} & {$4.46$} & $24.92$ \\
    MIRT~\citep{song2025irt_router}                                           & {$46.84$} & {$28.25$} & {$16.28$} & {$4.48$} & $23.96$ \\
    IRT-Router~\citep{song2025irt_router}          & {$43.67$} & {$26.82$} & {$15.62$} & {$4.43$} & $22.64$ \\
    \midrule
    Top-5 router spread (max$-$min) & 0.33 & 0.20 & 0.10 & 0.10 & 0.18 \\
    \bottomrule
  \end{tabular}
  \vspace{-3mm}
\end{table}

\subsection{{The Plateau Persists Across Evaluation Settings}}
\label{sec:exp:robustness}

\paragraph{The plateau is stable across encoders.}
{We test whether the plateau is driven by a particular query representation.}
{Fig.~\ref{fig:encoder_spread} reports the standard deviation of test accuracy across encoders for each routing method. Averaged over the methods shown, this standard deviation ranges from $0.07$ to $0.65$~pp across benchmarks.}
{This suggests that the plateau is not an artifact of a specific encoder choice, and we include per-encoder breakdowns in App.~\ref{app:additional_results}.}

\paragraph{{The plateau persists across benchmark settings.}}
{We further test whether the plateau depends on the query distribution.
First, restricting the benchmarks to easier queries, which more models answer correctly, reduces the oracle gap but does not eliminate the plateau: leading routers still achieve similar accuracy far below the oracle.
Second, grouping \NBshortk{} queries by task domain shows that leading routers remain close in accuracy and substantially below the oracle within domains such as math and question answering.
Collectively, these observations suggest that the plateau is not specific to a particular benchmark or query mixture, but is a pattern shared by the routers we test.
App.~\ref{app:method-vs-benchmark} provides the experimental setups and detailed results.}

\paragraph{{Multi-sample correctness labels.}}
{We also test whether the plateau and oracle gap could arise from noise in single-sample correctness labels.}
{On \NBshortk{}, ten-sample majority labels yield an oracle accuracy of $0.817$, compared with $0.827$ under single-sample labels, retaining about $93\%$ of the gap to the original best-router accuracy.}
{These results suggest that neither the plateau nor the large oracle gap is primarily an artifact of noise in single-sample correctness labels.}
More detail in App.~\ref{app:multisample}.

\begin{figure}[t]
  \centering
  \includegraphics[width=0.65\textwidth]{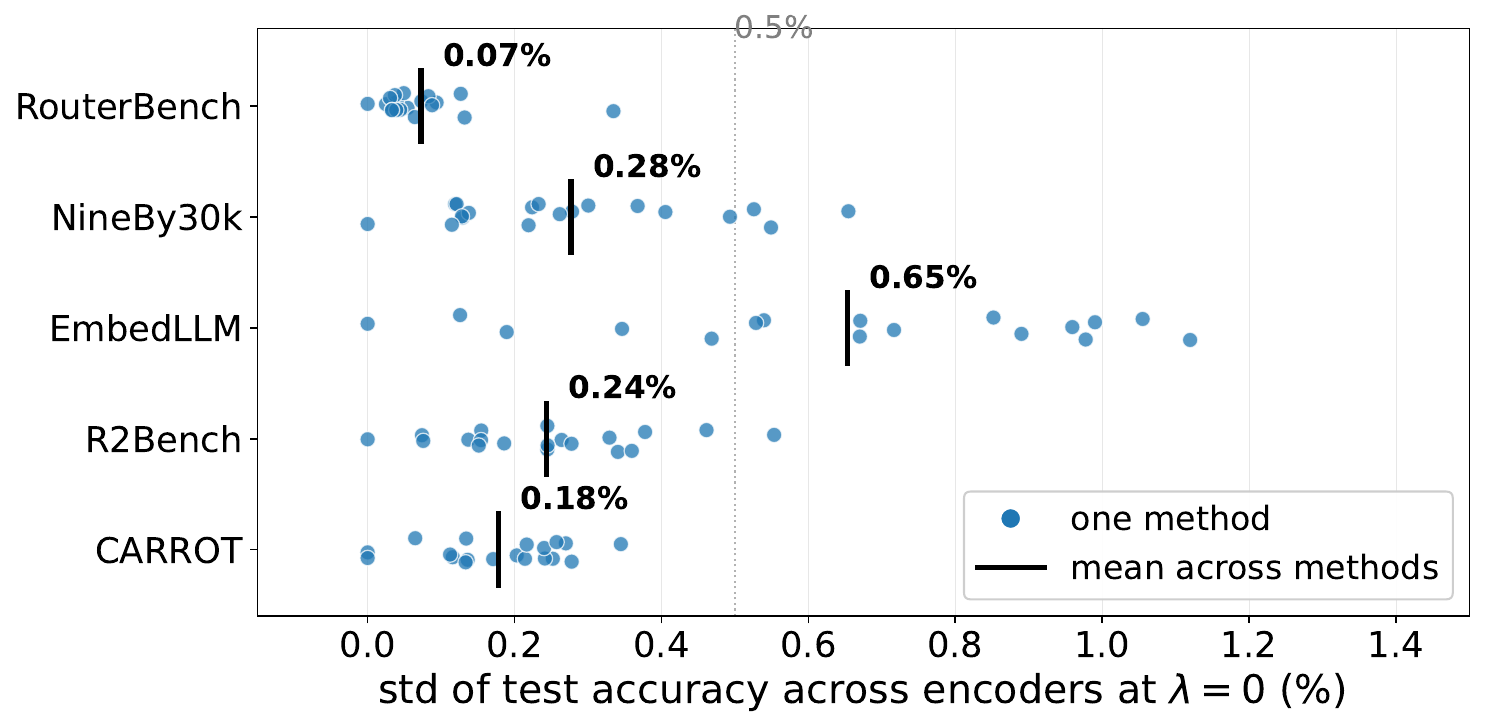}
  \vspace{-1mm}
  \caption{{Standard deviation of test accuracy across encoders at $\lambda{=}0$, computed per method and benchmark. Each blue dot is a routing method and black ticks show the mean across methods.}}
\vspace{-2mm}
  \label{fig:encoder_spread}
\end{figure}

\section{Understanding the Routing Plateau}
\label{sec:plateau-analysis}

We next seek to understand why the routing plateau occurs.
Our hypothesis is that it arises from a \textbf{correctness-prediction bottleneck}: before selecting a model, a router must infer how likely each candidate model can answer the query correctly.
However, current routers tend to learn coarse, global model-capability patterns rather than fine-grained, instance-level differences in correctness.
We support this hypothesis through the analysis in this section.

\subsection{Coarse-Grained Correctness Prediction Limits Routing Accuracy}
\label{sec:plateau-stratification}

\begin{figure}[t]
  \centering
  \includegraphics[width=1\textwidth]{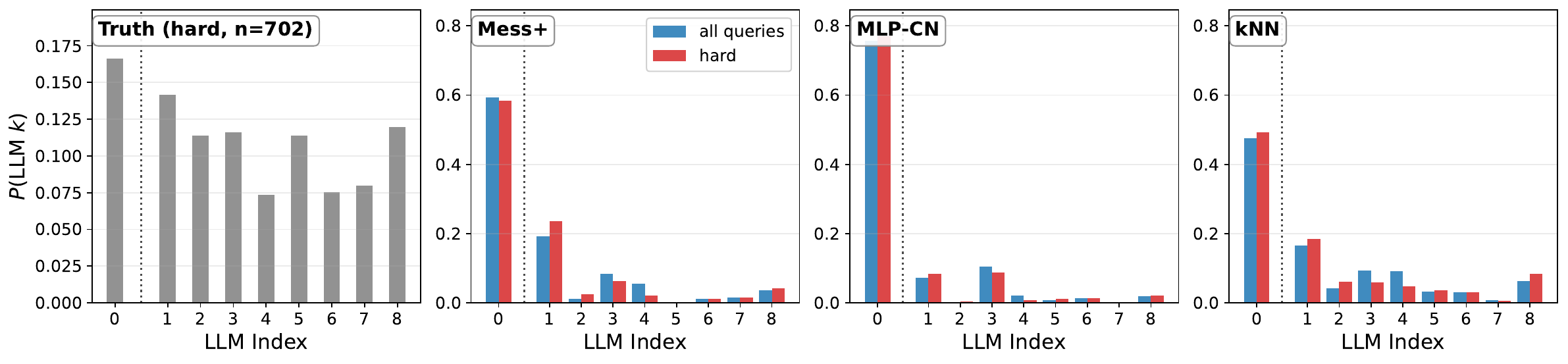}
  \caption{{Model selection on hard (red) and all (blue) \NBshortk{} queries. Routers favor globally strong models, while correct-model mass on hard queries (gray) is more evenly distributed.}}
\vspace{-1mm}
  \label{fig:score_spread_comparison}
\end{figure}

{To examine whether routers distinguish model correctness at the query level, we group queries by the number of models that answer them correctly.
Let $n_{\mathrm{correct}}(i)$ denote this number for query $q_i$.
We call a query \emph{hard} if $n_{\mathrm{correct}}(i) \le t$, where $t = \max(3, \lfloor 0.3K \rfloor)$, and \emph{easy} otherwise.
We could examine how routers choose models when correct answers are from only a small subset of the pool.}

{Fig.~\ref{fig:score_spread_comparison} compares model-selection frequencies on hard queries and on all queries for two learned routers and kNN, alongside the distribution of correct models on hard queries.}
{
First, model-selection frequencies change little between hard queries and the full query set.
Second, routers select the globally top-ranked model far more often than other models, whereas the ground-truth correctness distribution is more evenly spread across the pool.
This mismatch suggests that routers favor globally strong models rather than learning fine-grained, query-specific differences in model correctness.}
{We repeat this analysis across multiple routers and observe the same pattern in App.~\ref{app:more-examples}.}

\begin{table}[t]
\centering
\small
\caption{{Oracle--router gaps averaged over 12 routers. $f_{\text{hard}}$ is the hard-query fraction; Ratio is hard/easy gap; Gap share is hard queries' contribution to the total gap.}}
\label{tab:strata}
\setlength{\tabcolsep}{5pt}
\renewcommand{\arraystretch}{1.05}
\begin{tabular}{l c c c r r}
\toprule
Benchmark & $f_{\text{hard}}$ & {Hard gap} & {Easy gap} & Ratio & {Gap share} \\
\midrule
\RBench{}    & 0.132 & 0.515 & 0.046 & 11.3$\times$ & 64.1\% \\
\CarrotB{}   & 0.085 & 0.655 & 0.054 & 12.1$\times$ & 53.3\% \\
\RTwoBench{} & 0.160 & 0.523 & 0.049 & 10.6$\times$ & 68.0\% \\
\NBshortk{}  & 0.140 & 0.664 & 0.073 &  9.0$\times$ & 64.8\% \\
\EmbedLLM{}  & 0.432 & 0.570 & 0.108 &  5.3$\times$ & 80.7\% \\
\bottomrule
\vspace{-7mm}
\end{tabular}%
\end{table}

{We next quantify how much hard queries contribute to routing failures.
Tab.~\ref{tab:strata} shows that, on average across routers, the rate of selecting an incorrect model is $5.3$--$12.1$ times as high on hard queries as on easy ones.
Although hard queries are a minority of each benchmark, they account for $53.3\%$--$80.7\%$ of the overall oracle--router gap.
Together, these results suggest that routers capture global capability patterns but struggle to identify which models will succeed on individual hard queries.}

\subsection{Prediction Variation Across Routers Yields Error Exchange, Not Accuracy Gaps}
\label{sec:plateau-exchange}

Fig.~\ref{fig:score_spread_comparison} shows that routers can learn slightly different correctness-prediction functions and therefore select different models on some queries.
These differences can arise from their distinct inductive biases, training objectives, and optimization procedures.
However, Sec.~\ref{sec:experiments} shows that these differences do not yield clear aggregate accuracy gains: top routers still converge to the same plateau.
Our hypothesis is that router disagreement mostly produces \emph{error exchange} rather than consistent improvement.
{One router may solve queries another misses while failing on a comparable number that the other solves, so their wins and losses largely cancel out in overall accuracy.}

{To test this hypothesis, we consider each pair of routers $(r_a,r_b)$ and measure their aggregate accuracy gap, $|\mathrm{Acc}(r_a)-\mathrm{Acc}(r_b)|$, and their model-selection disagreement rate.}
{We further decompose disagreements into exchanged wins: queries solved by $r_a$ but missed by $r_b$, and vice versa.}
{App.~\ref{app:exchange-details} gives the computation details.}
Tab.~\ref{tab:exchange} shows that router differences are mostly balanced rather than directional.
{Top-tier routers can disagree on up to $70\%$ of queries, yet their median accuracy gaps remain small ($0.3$--$0.6$~pp) because their wins are nearly symmetric.}
{Conditioned on disagreement, the median unsigned outcome change is $|\Delta Y|{=}9.6\%$--$22.5\%$, while the signed change is $5$--$17{\times}$ smaller.}
Thus, prediction variation mainly induces error exchange rather than consistent accuracy separation, keeping routers clustered on the same plateau.

\begin{table}[t]
\centering
\small
\setlength{\tabcolsep}{4pt}
\caption{{Disagree is the model-selection disagreement range, and Acc. gap is the median accuracy difference. Within disagreements, $|\Delta Y|$ measures exchanged wins and $|\Delta Y_{\mathrm{signed}}|$ their net imbalance. Disagreement and unsigned exchange are in \%, accuracy gaps and signed exchange in pp.}}
\label{tab:exchange}
\resizebox{0.75\columnwidth}{!}{%
\begin{tabular}{lccccc}
\toprule
Benchmark & {Disagree range} & {Acc. gap} & $|\Delta Y|$ & $|\Delta Y_{\mathrm{signed}}|$ & $|\Delta Y|/|\Delta Y_{\mathrm{signed}}|$ \\
\midrule
\RBench{}     & {0.8--19.4} & {0.27} & {10.8} & {2.4} & 5$\times$ \\
\CarrotB{}    & {16.0--70.0} & {0.27} & {9.6} & {0.5} & 17$\times$ \\
\RTwoBench{}  & {14.9--44.2} & {0.27} & {16.2} & {1.0} & 17$\times$ \\
\NBshortk{}   & {3.2--52.7} & {0.46} & {14.5} & {1.1} & 13$\times$ \\
\EmbedLLM{}   & {16.0--53.0} & {0.55} & {22.5} & {1.8} & 13$\times$ \\
\bottomrule
\end{tabular}
\vspace{-10mm}
}
\end{table}

\subsection{{What Does Learning Add Beyond Local Similarity?}}
\label{sec:plateau-knn}

{We next ask whether learned routers predict model correctness better than kNN, which estimates each model's correctness from neighboring training queries in the embedding space.
We compare this local estimator with learned routers in routing accuracy and concordance, using the evaluated frozen query representations.
Concordance (Sec.~\ref{sec:prelim}) evaluates the ranking of correct models above incorrect ones, complementing routing accuracy.}
{App.~\ref{app:knn-comparison} details the comparison setup.}

kNN is among the leading methods on every benchmark (Tab.~\ref{tab:main:lambda0_all}).
Learned routers also show no clear advantage over kNN on hard queries, where fine-grained prediction of individual models' correctness is especially important (Tab.~\ref{tab:hard-acc}).
Their concordance is also similar to that of kNN's local similarity estimates, with gains over kNN limited to $0.006$ across the five benchmarks (Tab.~\ref{tab:knn-concordance}).

\begin{table}[t]
\centering
\small
\caption{kNN vs.\ trained routers. (a) Within-query ranking quality (concordance): trained routers have similar performance to kNN. (b) Hard-subset accuracy, $\Delta = \text{median trained} - \text{kNN}$: kNN matches the median trained routers.}

\label{tab:knn-vs-trained}
\begin{subtable}[t]{0.48\textwidth}
\centering
\subcaption{Test concordance: kNN is comparable with the trained routers.}
\label{tab:knn-concordance}
\setlength{\tabcolsep}{4pt}
\begin{tabular}{@{}lcc@{}}
\toprule
Benchmark & kNN & 11 other routers \\
\midrule
\RBench{}     & 0.768 & 0.762--0.772  \\
\CarrotB{}    & 0.805 & 0.794--0.804  \\
\RTwoBench{}  & 0.887 & 0.874--0.887  \\
\NBshortk{}   & 0.665 & 0.637--0.667  \\
\EmbedLLM{}   & 0.680 & 0.676--0.686  \\
\bottomrule
\end{tabular}%
\end{subtable}
\hfill
\begin{subtable}[t]{0.48\textwidth}
\centering
\subcaption{kNN matches the trained router on Hard-subset.}
\label{tab:hard-acc}
\vspace{\baselineskip}
\setlength{\tabcolsep}{4pt}
\begin{tabular}{@{}lcc@{}}
\toprule
Benchmark & kNN & Median Acc. $\Delta$ \\
\midrule
\RBench{}     & 0.478 & $+0.008$ \\
\CarrotB{}    & 0.331 & $+0.014$ \\
\RTwoBench{}  & 0.484 & $-0.005$ \\
\NBshortk{}   & 0.340 & $-0.007$ \\
\EmbedLLM{}   & 0.444 & $-0.015$ \\
\bottomrule
\end{tabular}%
\end{subtable}
\end{table}

{These results suggest that kNN provides a strong local approximation of model correctness from the evaluated query embeddings and training labels.
The learned routers offer little improvement over this local estimate, especially on hard queries.
This supports our correctness-prediction bottleneck hypothesis: methods share a difficulty in predicting instance-level correctness from frozen query representations.
Sec.~\ref{sec:fix} tests whether strengthening the training regime through more data, larger encoders, and end-to-end fine-tuning substantially changes this picture.}

\section{{Does a Stronger Training Regime Alleviate the Plateau?}}
\label{sec:fix}

{Finally, we ask whether the plateau reflects limitations of the training regime: insufficient data, limited encoder capacity, or query representations that are not adapted to routing.}
{If these constraints largely explain the plateau, a stronger training regime should substantially narrow the oracle gap.}
{Holding the router designs fixed, we study three changes: \emph{(1) scaling the training data, (2) using larger query encoders, and (3) fine-tuning the router end-to-end.}}

\subsection{Experimental Setup}
\label{sec:fix-setup}

\paragraph{{Datasets and encoders.}}
{To study data scaling, we use our paired benchmarks \NBshortk{} and \NBlongk{}, expanding the training set from $23$K to $300$K queries with correctness labels for the same nine candidate models. Both benchmarks share identical validation and test splits.}
{Construction details and exact split sizes are provided in App.~\ref{app:dataset}.}
{To study encoder scaling, we compare two encoders from the same architecture family: \textsc{ModernBERT-base} (\texttt{mbb}, ${\sim}110$M parameters) and \textsc{ModernBERT-large} (\texttt{mbl}, ${\sim}340$M parameters).}

\paragraph{Fine-tunable routing methods.}
{We evaluate eight routers from the main grid whose implementations support end-to-end encoder fine-tuning: \texttt{mlp\_cn}, \texttt{embedllm}, \texttt{routerdc}, \texttt{p2l}, \texttt{cp\_router}, \texttt{mess\_plus}, \texttt{zooter\_lite}, and \texttt{equi}.}
{We retain \texttt{kNN} as a frozen reference to measure the effect of data and encoder scaling without fine-tuning.}

\paragraph{{Configurations.}}
{We denote configurations by (data scale, training mode, encoder), with 30k and 300k referring to \NBshortk{} and \NBlongk{}, respectively.}
{The training mode either keeps the encoder frozen or fine-tunes it end-to-end (FT).}
{Our \emph{baseline} is (30k, frozen, mbb), and \emph{Scaled-FT} is (300k, FT, mbl).}
{Additional training settings are provided in App.~\ref{app:scaling-setup}.}

\subsection{{Scaling and Fine-Tuning Yield Modest Gains}}
\label{sec:fix-headline}

{Scaled-FT improves all eight evaluated routers, with gains of $0.31$--$2.13$~pp and a mean gain of $1.24$~pp (Fig.~\ref{fig:fix-headline}).
Concordance also improves: on the shared test split, the eight fine-tuned routers achieve a mean of $0.680$, compared with at most $0.667$ among the frozen routers evaluated in Sec.~\ref{sec:plateau-knn}.
However, seven of the eight routers still fall within a $0.85$~pp accuracy band, and their mean accuracy remains $13.27$~pp below the oracle.
The advantage over kNN also remains modest: under the same data and encoder scaling, frozen kNN improves by $0.49$~pp and matches the fine-tuned routers' mean accuracy of $69.38\%$.
Thus, strengthening the training regime improves accuracy and correctness ranking, but leaves routers closely clustered and far below the oracle.}

\begin{figure}[t]
\centering

\begin{minipage}[t]{0.47\linewidth}
\centering
\vspace{0pt}
\includegraphics[width=\linewidth]{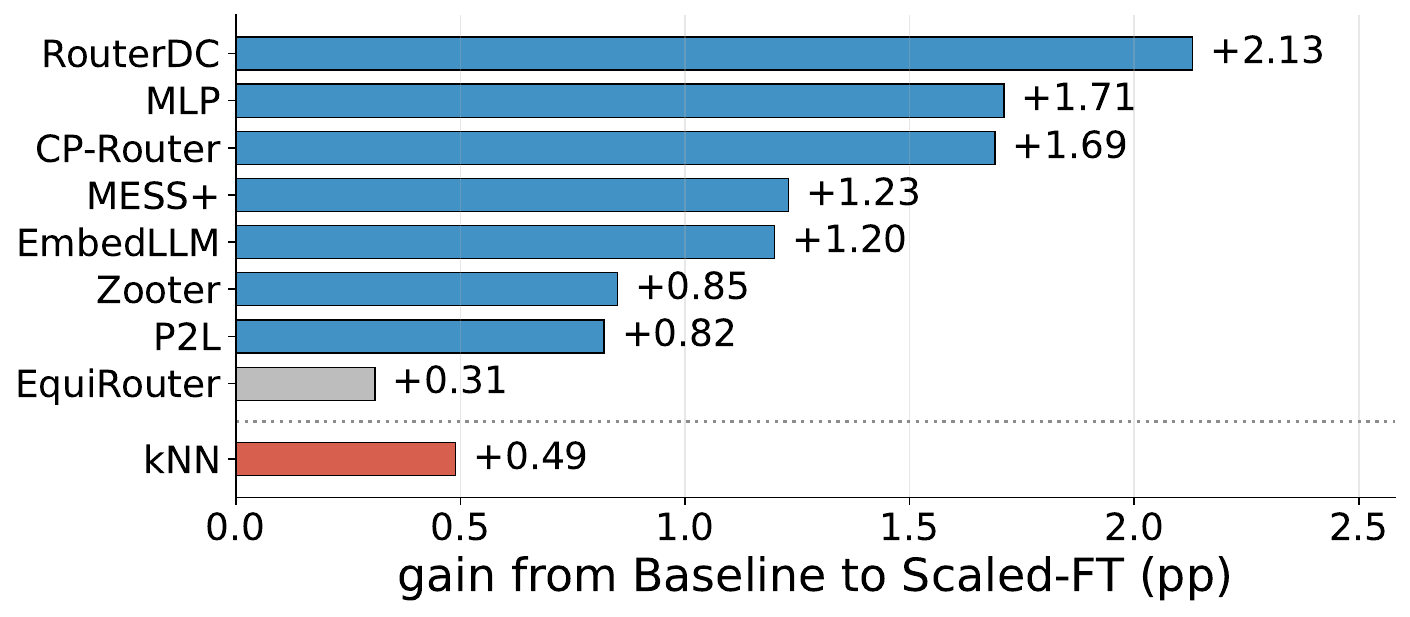}
\captionof{figure}{Per-method gain in test routing accuracy on \NBshortk{}$\to$\NBlongk{}, from Baseline (30k, frozen, mbb) to Scaled-FT (300k, FT, mbl). \texttt{kNN} has no fine-tuning axis and serves as the data-and-encoder-scaling-only comparison target.}
\label{fig:fix-headline}
\end{minipage}
\hfill
\begin{minipage}[t]{0.50\linewidth}
\centering
\vspace{0pt}
\captionof{table}{Ablation study of data scaling, encoder scaling, and router fine-tuning. Mean accuracy is reported across the eight trained methods. \emph{enc}: ModernBERT-large (mbl) vs.\ ModernBERT-base (mbb); \emph{data}: 300k vs.\ 30k; \emph{FT}: end-to-end fine-tuning vs.\ frozen encoder.}
\label{tab:fix-ablation}
\scriptsize
\setlength{\tabcolsep}{3.5pt}
\renewcommand{\arraystretch}{0.92}
\resizebox{\linewidth}{!}{
\begin{tabular}{@{}cccccc@{}}
\toprule
enc & data & FT & Cell & Mean acc & $\Delta$ pp \\
\midrule
            &              &              & 30k, frzn, mbb          & 0.6814          & ---       \\
$\checkmark$ &              &              & 30k, frzn, mbl          & 0.6827          & $+0.13$   \\
 & $\checkmark$ & & 300k, frzn, mbb     & 0.6861 & $+0.47$ \\
            &              & $\checkmark$ & 30k, FT, mbb            & 0.6856          & $+0.42$   \\
$\checkmark$ &              & $\checkmark$ & 30k, FT, mbl            & 0.6851          & $+0.37$   \\
            & $\checkmark$ & $\checkmark$ & 300k, FT, mbb           & 0.6933          & $+1.19$   \\
$\checkmark$ & $\checkmark$ & $\checkmark$ & 300k, FT, mbl (fix)     & \textbf{0.6938} & $\mathbf{+1.24}$ \\
\bottomrule
\end{tabular}
}
\end{minipage}
\end{figure}

\subsection{Ablation Study}
\label{sec:fix-ablation}

{Tab.~\ref{tab:fix-ablation} separates the contributions of data scaling, encoder scaling, and fine-tuning, averaging accuracy across the eight routers.
Each change alone improves accuracy by less than $0.5$~pp.
Combining more data with fine-tuning yields a larger gain of $1.19$~pp with the base encoder, while increasing encoder size adds only another $0.05$~pp.
Thus, most of the improvement comes from jointly expanding the training data and adapting the query representation, with little benefit from encoder scaling.}

{On \CarrotB{} and \EmbedLLM{}, which lack data-scaled variants, we evaluate encoder scaling and end-to-end fine-tuning with the original training sets.}
{Relative to each benchmark's frozen-baseline, the fine-tuned large encoder improves mean accuracy by $0.50$ and $1.64$~pp, respectively.}
{The full encoder-by-training-mode grids are reported in App.~\ref{app:fix-crossbench}.}

\subsection{{Discussion: The Remaining Oracle Gap}}
\label{sec:fix-discussion}

{Across the routing methods and query representations we evaluate, routers reach similar accuracy while leaving a substantial gap to the oracle.
More training data, larger encoders, and end-to-end fine-tuning yield modest gains but do not substantially change this pattern.
These interventions improve how routers learn from queries, while leaving their inputs at routing time unchanged: they must still predict correctness before observing any candidate model's response.
Attempting a task could reveal model-specific difficulties that are not apparent from the query's semantics alone.
This suggests that further progress may require enriching the inputs to routing with evidence from model execution, such as partial output trajectories or confidence estimates derived from an initial attempt.}

\section{Conclusion}
\label{sec:conclusion}

Our extensive study of 21 routing methods across five representative benchmarks reveals the \emph{routing plateau}: many routers, including kNN-style methods, cluster around a similar accuracy ceiling that remains far below the oracle router.
Our analysis supports a correctness-prediction bottleneck hypothesis: current routers largely capture coarse, global patterns of model capability, but struggle to identify which models will succeed on individual hard queries.
Using our \NBlongk{} benchmark, we find that more training data, larger encoders, and end-to-end fine-tuning improve accuracy by $1.24$~pp on average across eight routers, while leaving the plateau largely intact.
These findings suggest that approaching the oracle may require richer routing inputs beyond the query alone.

\bibliographystyle{iclr2027_conference}
\bibliography{references}

\clearpage
\appendix
\section{Experimental Setup}

\subsection{Router Training and Evaluation}
\label{app:experiments}

\subsubsection{Benchmarks and preprocessing}
\label{app:experiments:preprocessing}
Tab.~\ref{tab:benchmarks} summarizes the available benchmark splits. 
The main comparison uses shared evaluation queries across routers and encoders: 10{,}950 on \RBench{}, 5{,}000 on \NBshortk{}, 3{,}000 on \EmbedLLM{}, 9{,}291 on \RTwoBench{}, and 6{,}637 on \CarrotB{}. 
Query embeddings are cached once per benchmark and encoder and reused by all frozen-encoder methods.

\begin{table}[t]
  \centering
  \small
  \setlength{\tabcolsep}{6pt}
  \caption{Benchmark suite. Train/Val/Test are available unique-query counts; $K$ is the pool size; Cost marks cost availability. For \EmbedLLM{}, cost is approximated by model parameter count. The main grid evaluates 3{,}000 \EmbedLLM{} and 6{,}637 \CarrotB{} test queries; the table lists their full available test splits.}
  \begin{tabular}{lrrrrc}
    \toprule
    Benchmark & Train & Val & Test & $K$ & Cost \\
    \midrule
    \NBshortk{} (ours)         &  23{,}027 &  3{,}000 &  5{,}000 &   9 & \checkmark \\
    \NBlongk{} (ours)          & 302{,}371 &  3{,}000 &  5{,}000 &   9 & \checkmark \\
    \RBench{}                  &  21{,}898 &  3{,}649 & 10{,}950 &  11 & \checkmark \\
    \RTwoBench{}               &  18{,}580 &  3{,}097 &  9{,}291 &  10 & \checkmark \\
    \EmbedLLM{}                & 119{,}964 & 14{,}897 & 15{,}210 & 112 & Proxy \\
    \CarrotB{}                 & 152{,}193 & 35{,}528 & 36{,}214 &  13 & \checkmark \\
    \bottomrule
  \end{tabular}
  \label{tab:benchmarks}
\end{table}

The benchmarks provide task scores $s_{im}\in[0,1]$ for query $i$ and candidate model $m$; some tasks give partial credit. Routing accuracy in the main grid is the mean score of the selected model, and the oracle averages the maximum score within each query. The behavioral analyses in Sec.~\ref{sec:plateau-stratification} instead use binary labels $y_{im}=\mathbf{1}[s_{im}\ge 0.5]$, as detailed in App.~\ref{app:more-examples}. All methods in a comparison use the same score matrix and model ordering. Cost annotations are used only for the cost-utility analysis at $\lambda>0$.

\subsubsection{Training, selection, and averaging}
\label{app:experiments:protocol}
Hyperparameter selection uses two stages. For each benchmark, method, and encoder, we first evaluate up to 15 architectures with seed 42 and retain the top five by validation routing accuracy at $\lambda=0$. These architectures are then evaluated with four additional seeds, $\{1337,2026,7,99\}$. The result aggregation first prefers architectures with the most completed seeds and then selects the highest mean validation accuracy. Trained methods in the main grid use the checkpoint with the best validation routing accuracy.

Training objectives follow the router implementation: classifier-style heads use BCE-based losses, distillation methods combine KL and BCE terms, P2L uses a Bradley--Terry objective, and latent-factor methods use their corresponding objectives. The cost sweep applies $U(\lambda)=(1-\lambda)\hat{s}-\lambda\tilde c$ after training, for $\lambda\in\{0,0.2,0.5,0.7,0.9\}$, retaining the architecture selected at $\lambda=0$.

\subsubsection{Hardware}
\label{app:experiments:hardware}
The frozen-encoder experiments used NVIDIA RTX A5000 GPUs (24\,GB) and NVIDIA L40S GPUs (48\,GB), on nodes with eight GPUs each.

\subsection{Dataset Construction and Splits}
\label{app:dataset}

This appendix section documents the construction, splits, and scope of the two paired benchmarks introduced in Sec.~\ref{sec:exp:benchmarks}.

\subsubsection{Construction}
\label{app:dataset:construction}

\paragraph{Source of the prompts.}
Prompts are aggregated from 57 datasets covering knowledge, reasoning, code, commonsense, safety, and multilingual QA. Fifty-six are publicly released benchmarks; one (PANDAX-patent-QA) is an internal patent question-answering split included for domain coverage. Each prompt is rewritten in a uniform format with an explicit answer-extraction instruction so that responses can be passed to the correct task-specific scorer. Per-source query counts are summarized in Tab.~\ref{tab:dataset:per-source-counts}.
The full source list, grouped by domain, is:
\begin{itemize}\setlength{\itemsep}{2pt}\setlength{\parsep}{0pt}
  \item \emph{Multiple-choice knowledge, commonsense, and safety:} MMLU~\citep{hendrycks2021mmlu}, MMLU-Pro~\citep{wang2024mmlupro}, ARC-Challenge and ARC-Easy~\citep{clark2018arc}, ArcMMLU\footnote{\url{https://huggingface.co/datasets/patrickshitou/ArcMMLU}}, BoolQ~\citep{clark2019boolq}, MedQA~\citep{jin2021medqa}, MedMCQA~\citep{pal2022medmcqa}, GPQA~\citep{rein2023gpqa}, SciQ~\citep{welbl2017sciq}, OpenBookQA~\citep{mihaylov2018openbookqa}, CommonsenseQA~\citep{talmor2019commonsenseqa}, HellaSwag~\citep{zellers2019hellaswag}, Winogrande~\citep{sakaguchi2020winogrande}, TruthfulQA~\citep{lin2022truthfulqa}, ETHICS~\citep{hendrycks2021ethics}, QASC~\citep{khot2020qasc}, BioASQ~\citep{tsatsaronis2015bioasq}, FinQA~\citep{chen2021finqa}, CaseHOLD~\citep{zheng2021casehold}, MusicTheoryBench~\citep{yuan2024chatmusician}.
  \item \emph{Reading-comprehension and open-domain QA:} SQuAD~\citep{rajpurkar2016squad}, NQ-Open~\citep{kwiatkowski2019nq}, HotpotQA~\citep{yang2018hotpotqa}, NarrativeQA~\citep{kocisky2018narrativeqa}, CoQA~\citep{reddy2019coqa}, QuAC~\citep{choi2018quac}, NewsQA~\citep{trischler2017newsqa}, SearchQA~\citep{dunn2017searchqa}, MS~MARCO~\citep{bajaj2018msmarco}, WebQuestions~\citep{berant2013webquestions}, WikiHop~\citep{welbl2018wikihop}, MLQA~\citep{lewis2020mlqa}, TyDi~QA~\citep{clark2020tydi}, XQuAD~\citep{artetxe2020xquad}, Social~IQa~\citep{sap2019socialiqa}, Qanta~\citep{rodriguez2019qanta}, ChemistryQA\footnote{\url{https://huggingface.co/datasets/avaliev/ChemistryQA}}, PANDAX-patent-QA\footnote{Custom internal split of a patent question-answering corpus.}.
  \item \emph{Math and quantitative reasoning:} GSM8K~\citep{cobbe2021gsm8k}, MATH~\citep{hendrycks2021math}, MathQA~\citep{amini2019mathqa}, AQuA-RAT~\citep{ling2017aqua}, SVAMP~\citep{patel2021svamp}, AsDiv~\citep{miao2020asdiv}.
  \item \emph{Code generation:} HumanEval~\citep{chen2021humaneval}, MBPP~\citep{austin2021mbpp}, LiveCodeBench~\citep{jain2024livecodebench}.
  \item \emph{SuperGLUE-style tasks~\citep{wang2019superglue}:} causal reasoning, cloze test, textual entailment, multiple-choice QA, reading comprehension, words-in-context (WiC), and Winograd schema (WSC).
  \item \emph{Domain-specific:} LSAT-AR via AGIEval~\citep{zhong2023agieval}, ChessInstruct\footnote{\url{https://huggingface.co/datasets/Thytu/ChessInstruct}}.
\end{itemize}

\paragraph{Models.}
We evaluate the same nine candidate LLMs on every prompt: \texttt{deepseek-v3.2}, \texttt{gemma-3n-e4b-it}, \texttt{llama-3.3-70b-instruct}, \texttt{minimax-m2.5}, \texttt{glm-4.5-air}, \texttt{qwen3-235b-a22b-2507}, \texttt{qwen3-coder-next}, \texttt{grok-4.1-fast}, and \texttt{mimo-v2-flash}.\footnote{Names follow OpenRouter's provider-qualified identifiers; decoding settings match each provider's defaults at the time of evaluation.} The pool is intentionally heterogeneous in scale and family: a small open-weights model (\texttt{gemma-3n-e4b-it}, $\sim$4B), several mid-to-large dense and mixture-of-experts generalists, and one code-specialist (\texttt{qwen3-coder-next}). No single model dominates across task categories, so routing decisions are non-trivial throughout the pool.

\paragraph{Inference protocol.}
All nine models are queried through the OpenRouter API. Each call records the single-sample generated answer, prompt and completion token counts, the serving provider, and a success flag. Per-call cost is computed from the recorded token counts and each provider's posted per-token rates, giving a real-valued cost annotation for every (query, model) pair.

\paragraph{Correctness labels.}
Each (query, model) pair receives a task-specific score $s_{im}\in[0,1]$. These scores can be fractional; we retain them for the main routing-accuracy comparisons and threshold them at $0.5$ for the binary analyses. Multiple-choice tasks use exact letter-choice matching against a boxed answer; extractive QA uses exact match or token-level F1; math benchmarks use numeric or symbolic equivalence; code benchmarks use unit-test execution; NarrativeQA uses METEOR; SuperGLUE subtasks use their canonical metrics. For a small set of open-ended tasks where exact match is too brittle (QANTA, WebQuestions, ChemistryQA), we fall back to LLM-as-judge with \texttt{gpt-4o-mini}. The rule-based path is always tried first and the judge is only invoked when it fails, so judge calls cover only a small fraction of the pool.

\begin{table}[H]
\centering
\footnotesize
\setlength{\tabcolsep}{4pt}
\renewcommand{\arraystretch}{0.95}
\caption{Per-source query counts in \NBshortk{} and \NBlongk{}. Counts are unique prompts. Sources are grouped by domain and sorted by \NBshortk{} count within each group. Total: \textbf{31{,}027} prompts in \NBshortk{}, \textbf{310{,}371} in \NBlongk{}.}
\label{tab:dataset:per-source-counts}
\begin{minipage}[t]{0.49\linewidth}
\centering
\begin{tabular}{@{}lrr@{}}
\toprule
Source & \NBshortk{} & \NBlongk{} \\
\midrule
\multicolumn{3}{@{}l}{\emph{MCQ knowledge / commonsense / safety}} \\
MMLU                  & 1{,}323 & 13{,}766 \\
ETHICS                & 1{,}127 & 11{,}170 \\
MMLU-Pro              & 1{,}100 & 11{,}108 \\
HellaSwag             & 1{,}024 & 10{,}042 \\
MedMCQA               & 626     & 6{,}149  \\
CaseHOLD              & 536     & 5{,}314  \\
BoolQ                 & 341     & 3{,}270  \\
ARC-Easy              & 244     & 2{,}376  \\
MedQA                 & 143     & 1{,}273  \\
CommonsenseQA         & 123     & 1{,}221  \\
Winogrande            & 121     & 1{,}267  \\
ARC-Challenge         & 114     & 1{,}172  \\
SciQ                  & 107     & 1{,}000  \\
QASC                  & 98      & 926      \\
TruthfulQA            & 82      & 817      \\
ArcMMLU               & 69      & 733      \\
FinQA                 & 69      & 670      \\
OpenBookQA            & 55      & 500      \\
GPQA                  & 52      & 546      \\
BioASQ                & 38      & 324      \\
MusicTheoryBench      & 33      & 367      \\
\midrule
\multicolumn{3}{@{}l}{\emph{Math and quantitative reasoning}} \\
MathQA                & 282     & 2{,}983  \\
AsDiv                 & 246     & 2{,}147  \\
GSM8K                 & 148     & 1{,}319  \\
MATH                  & 51      & 500      \\
AQuA-RAT              & 28      & 254      \\
SVAMP                 & 28      & 251      \\
\midrule
\multicolumn{3}{@{}l}{\emph{Code generation}} \\
MBPP                  & 52      & 499      \\
LiveCodeBench         & 50      & 511      \\
HumanEval             & 19      & 164      \\
\bottomrule
\end{tabular}
\end{minipage}\hfill
\begin{minipage}[t]{0.49\linewidth}
\centering
\begin{tabular}{@{}lrr@{}}
\toprule
Source & \NBshortk{} & \NBlongk{} \\
\midrule
\multicolumn{3}{@{}l}{\emph{Reading-comprehension and open-domain QA}} \\
SearchQA              & 4{,}404 & 43{,}216 \\
MLQA                  & 4{,}157 & 42{,}225 \\
Social IQa            & 3{,}322 & 33{,}375 \\
XQuAD                 & 1{,}458 & 14{,}196 \\
SQuAD                 & 1{,}075 & 10{,}532 \\
NarrativeQA           & 1{,}005 & 9{,}897  \\
MS MARCO              & 870     & 9{,}399  \\
HotpotQA              & 773     & 7{,}405  \\
CoQA                  & 726     & 7{,}242  \\
QuAC                  & 642     & 6{,}612  \\
WikiHop               & 525     & 5{,}128  \\
TyDi QA               & 494     & 5{,}077  \\
NewsQA                & 369     & 3{,}862  \\
NQ-Open               & 300     & 2{,}889  \\
WebQuestions          & 202     & 2{,}032  \\
Qanta                 & 197     & 1{,}900  \\
PANDAX-patent-QA      & 181     & 1{,}810  \\
ChemistryQA           & 50      & 494      \\
\midrule
\multicolumn{3}{@{}l}{\emph{SuperGLUE-style tasks}} \\
ClozeTest             & 935     & 9{,}999  \\
RC                    & 463     & 4{,}847  \\
QA                    & 319     & 3{,}270  \\
WiC                   & 69      & 638      \\
Entailment            & 25      & 333      \\
WSC                   & 14      & 95       \\
CausalReasoning       & 8       & 100      \\
\midrule
\multicolumn{3}{@{}l}{\emph{Domain-specific}} \\
LSAT-AR (AGIEval)     & 102     & 1{,}009  \\
ChessInstruct         & 13      & 150      \\
\bottomrule
\end{tabular}
\end{minipage}
\end{table}

\subsubsection{Splits and Scale}
\label{app:dataset:splits}

The split is built once and shared across every method we report, so cross-method gaps in Sec.~\ref{sec:exp:results} cannot be a split artefact.
We first take a uniform 10\% subsample of the source prompt pool; \NBshortk{} lives entirely inside this slice.
From it, we hold out 5{,}000 prompts as test and 3{,}000 as validation, leaving 23{,}027 prompts as the training set.
Every model is evaluated on every prompt in \NBshortk{}, so coverage is fully dense.

\NBlongk{} reuses \NBshortk{}'s test and validation splits without modification and replaces the training set with 302{,}371 prompts from the expanded source pool, excluding the held-out validation and test queries.
With router and encoder settings held fixed, the shared validation and test splits let us isolate the effect of training-data scale.
Coverage on the extension is dense: every prompt is scored by all nine models, giving $9 \times 310{,}371 = 2{,}793{,}339$ query--model correctness labels in \NBlongk{} overall.

\NBshortk{} is the basis for the main method grid (Sec.~\ref{sec:exp:benchmarks}); \NBlongk{} is the basis for the fine-tuning analysis (Sec.~\ref{sec:fix}).

\subsubsection{Position Relative to Existing Benchmarks}
\label{app:dataset:comparison}

Tab.~\ref{tab:benchmarks} places \NBshortk{} and \NBlongk{} alongside \RBench{}, \EmbedLLM{}, \RTwoBench{}, and \CarrotB{}, the four routing benchmarks we treat as prior art. The ``queries'' columns count unique prompts, not (prompt, model) entries. \NBshortk{} sits in the same scale band as the four existing benchmarks and serves as the apples-to-apples comparison set throughout this paper. \NBlongk{} provides a larger training set with the same candidate models and held-out queries, allowing us to vary the data scale without changing the evaluation setting.

\FloatBarrier

\section{Additional Experimental Results}

\label{app:robustness}

\subsection{Statistical Comparisons of the Top-Five Routers}
\label{app:significance}

We test whether the small accuracy differences among the top-five methods in Tab.~\ref{tab:main:lambda0_all} are statistically detectable.
For each benchmark, we identify the five highest-scoring methods using their encoder-averaged test accuracies.

\paragraph{Paired bootstrap.}
For routers $A$ and $B$ and test query $q_i$, let $d_i=s_A(q_i)-s_B(q_i)$, where each router's per-query score is averaged over the same encoders and seeds as its table entry.
We resample the $N$ test queries with replacement $10{,}000$ times and recompute the mean difference to obtain a $95\%$ confidence interval and a two-sided bootstrap $p$-value.
The comparison is paired because both routers are evaluated on the same queries, and it applies to both binary and fractional correctness scores.
We apply Holm's correction at $\alpha=0.05$ across the available pairwise comparisons among the top five on each benchmark.

\paragraph{Coverage and results.}
The test requires saved per-query predictions for every run contributing to a method's table entry; incomplete predictions limit coverage on four benchmarks.
Tab.~\ref{tab:app:top5-bootstrap} summarizes coverage and the largest observed gap on each benchmark. None of the top-five comparisons is significant after Holm correction.

\begin{table}[h]
\centering
\small
\setlength{\tabcolsep}{4pt}
\caption{Top-five bootstrap comparisons. Each row shows the tested pair with the largest observed accuracy gap. Confidence intervals are unadjusted, and $p$-values are Holm-adjusted over all tested pairs.}
\label{tab:app:top5-bootstrap}
\begin{tabular}{@{}llrr@{}}
\toprule
Benchmark & Pair ($A-B$) & Gap [95\% CI] (pp) & $p_{\mathrm{Holm}}$ \\
\midrule
RouterBench  & kNN--KMeans & $0.036\;[-0.039,\,0.112]$ & 1.000 \\
Nine-by-30k  & Zooter--Elo & $0.038\;[-0.289,\,0.370]$ & 1.000 \\
R2-Bench & Zooter--Avengers-Pro & $0.226\;[0.028,\,0.419]$ & 0.232 \\
CARROT/SPROUT & Elo--KMeans & $0.210\;[0.021,\,0.404]$ & 0.172 \\
EmbedLLM  & Zooter--Elo & $0.492\;[-0.117,\,1.109]$ & 0.350 \\
\bottomrule
\end{tabular}
\end{table}

\subsection{Per-Benchmark Cost-Utility Tables}
\label{app:additional_results:per_benchmark}

The four tables below report realized cost-utility on \RBench{},
\EmbedLLM{}, \RTwoBench{}, and \CarrotB{} under the same selection
protocol as Tab.~\ref{tab:main:nb30k_utility} in the main paper:
$16$ calibrated methods ($15$ available on \EmbedLLM{}), val-selected architecture per encoder,
mean over available encoders with cross-encoder std reported as $\pm$.
Costs are normalized by the maximum candidate cost for each query; utility values below are unscaled, whereas Tab.~\ref{tab:main:nb30k_utility} reports $100U$.
Methods are sorted by $\lambda{=}0.5$ utility within each table.

\begin{table}[H]
  \centering
  \small
  \setlength{\tabcolsep}{2.5pt}
  \caption{Realized cost-utility on \RBench{}. Same protocol and column meaning as Tab.~\ref{tab:main:nb30k_utility}; methods sorted by $\lambda{=}0.5$.}
  \label{tab:app:rb_utility}
  \begin{tabular}{lccccc}
    \toprule
     & $\lambda{=}0.0$ & $\lambda{=}0.2$ & $\lambda{=}0.5$ & $\lambda{=}0.7$ & $\lambda{=}0.9$ \\
    \midrule
    \texttt{kmeans}        & $0.8035\!\pm\!0.000$ & $\mathbf{0.5787}\!\pm\!0.001$ & $\mathbf{0.3333}\!\pm\!0.001$ & $0.1708\!\pm\!0.001$ & $0.0340\!\pm\!0.000$ \\
    \texttt{zooter\_lite}  & $0.8038\!\pm\!0.000$ & $0.5770\!\pm\!0.001$ & $0.3269\!\pm\!0.001$ & $0.1475\!\pm\!0.001$ & $0.0339\!\pm\!0.000$ \\
    \texttt{p2l}           & $0.8031\!\pm\!0.000$ & $0.5768\!\pm\!0.001$ & $0.3265\!\pm\!0.001$ & $0.1453\!\pm\!0.002$ & $0.0337\!\pm\!0.000$ \\
    \texttt{mess\_plus}    & $0.8035\!\pm\!0.000$ & $0.5740\!\pm\!0.003$ & $0.3261\!\pm\!0.001$ & $0.1476\!\pm\!0.002$ & $0.0339\!\pm\!0.000$ \\
    \texttt{embedllm}      & $0.8032\!\pm\!0.000$ & $0.5743\!\pm\!0.005$ & $0.3259\!\pm\!0.001$ & $0.1443\!\pm\!0.002$ & $0.0337\!\pm\!0.000$ \\
    \texttt{knn}           & $\mathbf{0.8039}\!\pm\!0.001$ & $0.5602\!\pm\!0.009$ & $0.3256\!\pm\!0.008$ & $\mathbf{0.1716}\!\pm\!0.001$ & $\mathbf{0.0341}\!\pm\!0.000$ \\
    \texttt{equi}          & $0.8036\!\pm\!0.000$ & $0.5717\!\pm\!0.002$ & $0.3254\!\pm\!0.001$ & $0.1458\!\pm\!0.001$ & $0.0337\!\pm\!0.000$ \\
    \texttt{mlp\_cn}       & $0.8029\!\pm\!0.000$ & $0.5720\!\pm\!0.003$ & $0.3254\!\pm\!0.001$ & $0.1485\!\pm\!0.003$ & $0.0338\!\pm\!0.000$ \\
    \texttt{avengers\_pro} & $0.8035\!\pm\!0.001$ & $0.5764\!\pm\!0.001$ & $0.3055\!\pm\!0.002$ & $0.1401\!\pm\!0.000$ & $0.0336\!\pm\!0.000$ \\
    \texttt{rorf}          & $0.7757\!\pm\!0.001$ & $0.5442\!\pm\!0.001$ & $0.3001\!\pm\!0.001$ & $0.1375\!\pm\!0.000$ & $0.0335\!\pm\!0.000$ \\
    \texttt{optllm\_xgb}   & $0.7668\!\pm\!0.003$ & $0.5359\!\pm\!0.005$ & $0.2982\!\pm\!0.001$ & $0.1373\!\pm\!0.000$ & $0.0335\!\pm\!0.000$ \\
    \texttt{carrot}        & $0.8030\!\pm\!0.001$ & $0.5728\!\pm\!0.007$ & $0.2967\!\pm\!0.013$ & $0.1414\!\pm\!0.001$ & $0.0336\!\pm\!0.000$ \\
    \texttt{elo\_router}   & $0.8036\!\pm\!0.001$ & $0.5716\!\pm\!0.007$ & $0.2896\!\pm\!0.013$ & $0.1408\!\pm\!0.001$ & $0.0336\!\pm\!0.000$ \\
    \texttt{mirt}          & $0.7636\!\pm\!0.059$ & $0.4472\!\pm\!0.044$ & $0.2473\!\pm\!0.009$ & $0.1375\!\pm\!0.002$ & $0.0335\!\pm\!0.000$ \\
    \texttt{cp\_router}    & $0.8029\!\pm\!0.001$ & $0.4123\!\pm\!0.001$ & $0.2401\!\pm\!0.001$ & $0.1361\!\pm\!0.000$ & $0.0335\!\pm\!0.000$ \\
    \texttt{irt\_router}   & $0.7749\!\pm\!0.062$ & $0.3917\!\pm\!0.000$ & $0.2381\!\pm\!0.000$ & $0.1358\!\pm\!0.000$ & $0.0335\!\pm\!0.000$ \\
    \bottomrule
  \end{tabular}
\end{table}

\begin{table}[H]
  \centering
  \small
  \setlength{\tabcolsep}{2.5pt}
  \caption{Realized cost-utility on \EmbedLLM{}, with model parameter count used as a cost proxy. \texttt{rorf} has only one viable encoder; cross-encoder std is therefore omitted for that row.}
  \label{tab:app:embedllm_utility}
  \begin{tabular}{lccccc}
    \toprule
     & $\lambda{=}0.0$ & $\lambda{=}0.2$ & $\lambda{=}0.5$ & $\lambda{=}0.7$ & $\lambda{=}0.9$ \\
    \midrule
    \texttt{kmeans}        & $0.6667\!\pm\!0.009$ & $0.4542\!\pm\!0.004$ & $\mathbf{0.2292}\!\pm\!0.003$ & $0.1114\!\pm\!0.002$ & $0.0219\!\pm\!0.001$ \\
    \texttt{knn}           & $0.6716\!\pm\!0.009$ & $\mathbf{0.4543}\!\pm\!0.008$ & $0.2281\!\pm\!0.004$ & $\mathbf{0.1120}\!\pm\!0.003$ & $\mathbf{0.0225}\!\pm\!0.001$ \\
    \texttt{zooter\_lite}  & $\mathbf{0.6751}\!\pm\!0.003$ & $0.4403\!\pm\!0.004$ & $0.2231\!\pm\!0.003$ & $0.1107\!\pm\!0.002$ & $0.0202\!\pm\!0.000$ \\
    \texttt{mess\_plus}    & $0.6699\!\pm\!0.005$ & $0.4388\!\pm\!0.005$ & $0.2218\!\pm\!0.003$ & $0.1092\!\pm\!0.001$ & $0.0202\!\pm\!0.001$ \\
    \texttt{embedllm}      & $0.6605\!\pm\!0.010$ & $0.4315\!\pm\!0.007$ & $0.2215\!\pm\!0.003$ & $0.1089\!\pm\!0.002$ & $0.0213\!\pm\!0.000$ \\
    \texttt{p2l}           & $0.6638\!\pm\!0.011$ & $0.4331\!\pm\!0.007$ & $0.2214\!\pm\!0.003$ & $0.1093\!\pm\!0.002$ & $0.0213\!\pm\!0.001$ \\
    \texttt{equi}          & $0.6708\!\pm\!0.005$ & $0.4378\!\pm\!0.003$ & $0.2208\!\pm\!0.003$ & $0.1062\!\pm\!0.002$ & $0.0198\!\pm\!0.001$ \\
    \texttt{mlp\_cn}       & $0.6651\!\pm\!0.007$ & $0.4298\!\pm\!0.008$ & $0.2200\!\pm\!0.003$ & $0.1085\!\pm\!0.002$ & $0.0212\!\pm\!0.000$ \\
    \texttt{carrot}        & $0.6698\!\pm\!0.011$ & $0.4173\!\pm\!0.007$ & $0.2019\!\pm\!0.007$ & $0.0897\!\pm\!0.001$ & $0.0191\!\pm\!0.000$ \\
    \texttt{elo\_router}   & $0.6702\!\pm\!0.007$ & $0.4176\!\pm\!0.006$ & $0.2018\!\pm\!0.006$ & $0.0892\!\pm\!0.002$ & $0.0191\!\pm\!0.000$ \\
    \texttt{avengers\_pro} & $0.6661\!\pm\!0.009$ & $0.4181\!\pm\!0.007$ & $0.1995\!\pm\!0.004$ & $0.0885\!\pm\!0.002$ & $0.0189\!\pm\!0.000$ \\
    \texttt{rorf}          & $0.6443$ & $0.4033$ & $0.1811$ & $0.0878$ & $0.0189$ \\
    \texttt{mirt}          & $0.5920\!\pm\!0.069$ & $0.2146\!\pm\!0.019$ & $0.1246\!\pm\!0.005$ & $0.0706\!\pm\!0.001$ & $0.0187\!\pm\!0.000$ \\
    \texttt{cp\_router}    & $0.6609\!\pm\!0.007$ & $0.2160\!\pm\!0.003$ & $0.1220\!\pm\!0.001$ & $0.0700\!\pm\!0.000$ & $0.0187\!\pm\!0.000$ \\
    \texttt{irt\_router}   & $0.5206\!\pm\!0.074$ & $0.1983\!\pm\!0.000$ & $0.1214\!\pm\!0.000$ & $0.0700\!\pm\!0.000$ & $0.0187\!\pm\!0.000$ \\
    \bottomrule
  \end{tabular}
\end{table}

\begin{table}[H]
  \centering
  \small
  \setlength{\tabcolsep}{2.5pt}
  \caption{Realized cost-utility on \RTwoBench{}, with token count used as a cost proxy. Negative values at $\lambda{=}0.9$ occur when the weighted normalized-cost term exceeds the weighted accuracy term.}
  \label{tab:app:r2bench_utility}
  \begin{tabular}{lccccc}
    \toprule
     & $\lambda{=}0.0$ & $\lambda{=}0.2$ & $\lambda{=}0.5$ & $\lambda{=}0.7$ & $\lambda{=}0.9$ \\
    \midrule
    \texttt{knn}           & $0.8208\!\pm\!0.002$ & $\mathbf{0.6139}\!\pm\!0.001$ & $\mathbf{0.3242}\!\pm\!0.001$ & $\mathbf{0.1475}\!\pm\!0.000$ & $\mathbf{-0.0082}\!\pm\!0.000$ \\
    \texttt{kmeans}        & $0.8187\!\pm\!0.002$ & $0.6125\!\pm\!0.001$ & $0.3232\!\pm\!0.001$ & $0.1467\!\pm\!0.000$ & $-0.0083\!\pm\!0.000$ \\
    \texttt{equi}          & $0.8180\!\pm\!0.003$ & $0.6060\!\pm\!0.001$ & $0.3139\!\pm\!0.001$ & $0.1437\!\pm\!0.000$ & $-0.0088\!\pm\!0.000$ \\
    \texttt{zooter\_lite}  & $\mathbf{0.8212}\!\pm\!0.001$ & $0.6069\!\pm\!0.001$ & $0.3132\!\pm\!0.001$ & $0.1434\!\pm\!0.000$ & $-0.0088\!\pm\!0.000$ \\
    \texttt{mlp\_cn}       & $0.8138\!\pm\!0.004$ & $0.6070\!\pm\!0.001$ & $0.3131\!\pm\!0.001$ & $0.1431\!\pm\!0.000$ & $-0.0088\!\pm\!0.000$ \\
    \texttt{mess\_plus}    & $0.8169\!\pm\!0.002$ & $0.6073\!\pm\!0.002$ & $0.3130\!\pm\!0.001$ & $0.1433\!\pm\!0.001$ & $-0.0088\!\pm\!0.000$ \\
    \texttt{p2l}           & $0.8156\!\pm\!0.003$ & $0.6048\!\pm\!0.002$ & $0.3094\!\pm\!0.001$ & $0.1416\!\pm\!0.001$ & $-0.0091\!\pm\!0.000$ \\
    \texttt{embedllm}      & $0.8140\!\pm\!0.005$ & $0.6035\!\pm\!0.003$ & $0.3087\!\pm\!0.003$ & $0.1412\!\pm\!0.001$ & $-0.0091\!\pm\!0.000$ \\
    \texttt{optllm\_xgb}   & $0.8043\!\pm\!0.003$ & $0.5953\!\pm\!0.002$ & $0.3064\!\pm\!0.000$ & $0.1369\!\pm\!0.000$ & $-0.0098\!\pm\!0.000$ \\
    \texttt{rorf}          & $0.8160\!\pm\!0.003$ & $0.6002\!\pm\!0.002$ & $0.3004\!\pm\!0.001$ & $0.1336\!\pm\!0.001$ & $-0.0100\!\pm\!0.000$ \\
    \texttt{carrot}        & $0.8154\!\pm\!0.002$ & $0.5923\!\pm\!0.001$ & $0.3001\!\pm\!0.000$ & $0.1342\!\pm\!0.000$ & $-0.0100\!\pm\!0.000$ \\
    \texttt{elo\_router}   & $0.8208\!\pm\!0.002$ & $0.5931\!\pm\!0.001$ & $0.2996\!\pm\!0.001$ & $0.1340\!\pm\!0.000$ & $-0.0101\!\pm\!0.000$ \\
    \texttt{avengers\_pro} & $0.8189\!\pm\!0.001$ & $0.5897\!\pm\!0.001$ & $0.2979\!\pm\!0.001$ & $0.1334\!\pm\!0.000$ & $-0.0101\!\pm\!0.000$ \\
    \texttt{mirt}          & $0.8094\!\pm\!0.006$ & $0.5295\!\pm\!0.068$ & $0.2785\!\pm\!0.025$ & $0.1295\!\pm\!0.010$ & $-0.0103\!\pm\!0.001$ \\
    \texttt{cp\_router}    & $0.8193\!\pm\!0.002$ & $0.5244\!\pm\!0.002$ & $0.2628\!\pm\!0.000$ & $0.1216\!\pm\!0.000$ & $-0.0113\!\pm\!0.000$ \\
    \texttt{irt\_router}   & $0.8104\!\pm\!0.005$ & $0.4442\!\pm\!0.003$ & $0.2473\!\pm\!0.001$ & $0.1177\!\pm\!0.000$ & $-0.0116\!\pm\!0.000$ \\
    \bottomrule
  \end{tabular}
\end{table}

\begin{table}[H]
  \centering
  \small
  \setlength{\tabcolsep}{2.5pt}
  \caption{Realized cost-utility on \CarrotB{}. Same protocol as Tab.~\ref{tab:main:nb30k_utility}.}
  \label{tab:app:carrot_utility}
  \begin{tabular}{lccccc}
    \toprule
     & $\lambda{=}0.0$ & $\lambda{=}0.2$ & $\lambda{=}0.5$ & $\lambda{=}0.7$ & $\lambda{=}0.9$ \\
    \midrule
    \texttt{knn}           & $0.8678\!\pm\!0.002$ & $\mathbf{0.6658}\!\pm\!0.002$ & $\mathbf{0.3948}\!\pm\!0.001$ & $\mathbf{0.2210}\!\pm\!0.000$ & $\mathbf{0.0625}\!\pm\!0.000$ \\
    \texttt{kmeans}        & $0.8664\!\pm\!0.001$ & $0.6641\!\pm\!0.002$ & $0.3945\!\pm\!0.001$ & $0.2201\!\pm\!0.000$ & $0.0622\!\pm\!0.000$ \\
    \texttt{zooter\_lite}  & $0.8664\!\pm\!0.001$ & $0.6612\!\pm\!0.001$ & $0.3891\!\pm\!0.001$ & $0.2151\!\pm\!0.001$ & $0.0613\!\pm\!0.000$ \\
    \texttt{mess\_plus}    & $0.8661\!\pm\!0.001$ & $0.6599\!\pm\!0.001$ & $0.3880\!\pm\!0.001$ & $0.2134\!\pm\!0.001$ & $0.0612\!\pm\!0.000$ \\
    \texttt{p2l}           & $0.8632\!\pm\!0.003$ & $0.6588\!\pm\!0.003$ & $0.3879\!\pm\!0.002$ & $0.2129\!\pm\!0.001$ & $0.0611\!\pm\!0.000$ \\
    \texttt{equi}          & $0.8646\!\pm\!0.002$ & $0.6580\!\pm\!0.002$ & $0.3875\!\pm\!0.002$ & $0.2145\!\pm\!0.001$ & $0.0613\!\pm\!0.000$ \\
    \texttt{mlp\_cn}       & $0.8612\!\pm\!0.002$ & $0.6558\!\pm\!0.003$ & $0.3875\!\pm\!0.002$ & $0.2131\!\pm\!0.001$ & $0.0611\!\pm\!0.000$ \\
    \texttt{embedllm}      & $0.8634\!\pm\!0.003$ & $0.6555\!\pm\!0.003$ & $0.3850\!\pm\!0.005$ & $0.2112\!\pm\!0.003$ & $0.0609\!\pm\!0.000$ \\
    \texttt{rorf}          & $0.8626\!\pm\!0.003$ & $0.6578\!\pm\!0.004$ & $0.3808\!\pm\!0.001$ & $0.2068\!\pm\!0.001$ & $0.0608\!\pm\!0.000$ \\
    \texttt{elo\_router}   & $\mathbf{0.8685}\!\pm\!0.002$ & $0.6594\!\pm\!0.002$ & $0.3767\!\pm\!0.002$ & $0.2054\!\pm\!0.001$ & $0.0602\!\pm\!0.000$ \\
    \texttt{carrot}        & $0.8637\!\pm\!0.003$ & $0.6574\!\pm\!0.002$ & $0.3759\!\pm\!0.002$ & $0.2060\!\pm\!0.001$ & $0.0601\!\pm\!0.000$ \\
    \texttt{optllm\_xgb}   & $0.8512\!\pm\!0.002$ & $0.6344\!\pm\!0.003$ & $0.3746\!\pm\!0.001$ & $0.2093\!\pm\!0.000$ & $0.0595\!\pm\!0.000$ \\
    \texttt{avengers\_pro} & $0.8674\!\pm\!0.001$ & $0.6581\!\pm\!0.001$ & $0.3738\!\pm\!0.001$ & $0.2041\!\pm\!0.001$ & $0.0599\!\pm\!0.000$ \\
    \texttt{cp\_router}    & $0.8665\!\pm\!0.003$ & $0.5754\!\pm\!0.006$ & $0.3246\!\pm\!0.001$ & $0.1873\!\pm\!0.000$ & $0.0568\!\pm\!0.000$ \\
    \texttt{mirt}          & $0.8488\!\pm\!0.012$ & $0.5369\!\pm\!0.085$ & $0.3193\!\pm\!0.040$ & $0.1838\!\pm\!0.017$ & $0.0561\!\pm\!0.003$ \\
    \texttt{irt\_router}   & $0.8478\!\pm\!0.010$ & $0.4714\!\pm\!0.016$ & $0.2886\!\pm\!0.005$ & $0.1705\!\pm\!0.002$ & $0.0535\!\pm\!0.000$ \\
    \bottomrule
  \end{tabular}
\end{table}

\FloatBarrier

\subsection{Per-Encoder Routing Accuracy}
\label{app:additional_results}
The following tables disaggregate Tab.~\ref{tab:main:lambda0_all} by query encoder. Each cell is the test accuracy of the validation-selected architecture. The final columns give the mean and population standard deviation across available encoders.

\begin{table}[htbp]
\centering
\small
\setlength{\tabcolsep}{4pt}
\caption{\RBench{}: per-encoder test routing accuracy (\%). Mean and SD summarize the encoder columns; SD is in percentage points.}
\label{tab:app:encoders:rb}
\begin{tabular}{@{}lrrrrrrrr@{}}
\toprule
Router & MiniLM & MPNet & Qwen & BGE & MBB & MBL & Mean & SD \\
\midrule
Zooter & 80.38 & 80.34 & 80.42 & 80.36 & 80.44 & 80.37 & 80.38 & 0.03 \\
kNN & 80.38 & 80.41 & 80.38 & 80.31 & 80.51 & 80.33 & 80.39 & 0.06 \\
Elo & 80.32 & 80.23 & 80.42 & 80.38 & 80.50 & 80.34 & 80.36 & 0.08 \\
Avengers-Pro & 80.34 & 80.37 & 80.31 & 80.25 & 80.43 & 80.34 & 80.34 & 0.05 \\
EquiRouter & 80.39 & 80.35 & 80.34 & 80.33 & 80.40 & 80.35 & 80.36 & 0.02 \\
KMeans & 80.30 & 80.31 & 80.41 & 80.31 & 80.42 & 80.35 & 80.35 & 0.05 \\
MESS+ & 80.29 & 80.29 & 80.39 & 80.37 & 80.40 & 80.35 & 80.35 & 0.04 \\
CARROT & 80.12 & 80.38 & 80.33 & 80.31 & 80.41 & 80.27 & 80.30 & 0.09 \\
CP-Router & 80.28 & 80.30 & 80.31 & 80.31 & 80.40 & 80.15 & 80.29 & 0.07 \\
P2L & 80.36 & 80.34 & 80.27 & 80.33 & 80.30 & 80.25 & 80.31 & 0.04 \\
MLP & 80.31 & 80.27 & 80.25 & 80.32 & 80.33 & 80.24 & 80.29 & 0.04 \\
EmbedLLM & 80.34 & 80.27 & 80.32 & 80.37 & 80.36 & 80.26 & 80.32 & 0.04 \\
ModelSpider & 80.31 & 80.30 & 80.36 & 80.24 & 80.32 & 80.30 & 80.31 & 0.03 \\
RoRF & 77.63 & 77.61 & 77.59 & 77.59 & 77.30 & 77.69 & 77.57 & 0.13 \\
TagRouter & 80.27 & 80.27 & 80.47 & 80.27 & 80.42 & --- & 80.34 & 0.09 \\
LinUCB & 80.29 & 80.23 & 79.16 & 80.31 & 55.80 & 64.71 & 73.41 & 9.66 \\
RouterDC & 80.20 & 79.97 & 80.03 & 80.18 & 80.33 & 80.31 & 80.17 & 0.13 \\
MIRT & 80.32 & 80.30 & 63.82 & 80.34 & 78.48 & 74.89 & 76.36 & 5.93 \\
IRT-Router & 80.27 & 80.27 & 63.61 & 80.27 & 80.27 & 80.25 & 77.49 & 6.21 \\
HybridLLM & 71.67 & 71.67 & 71.67 & 71.67 & 71.67 & 71.67 & 71.67 & 0.00 \\
OptLLM & 77.38 & 76.70 & 76.41 & 76.71 & 76.44 & 76.46 & 76.68 & 0.33 \\
\bottomrule
\end{tabular}

\end{table}

\begin{table}[htbp]
\centering
\small
\setlength{\tabcolsep}{4pt}
\caption{\NBshortk{}: per-encoder test routing accuracy (\%). Mean and SD summarize the encoder columns; SD is in percentage points.}
\label{tab:app:encoders:nb30k}
\begin{tabular}{@{}lrrrrrrrr@{}}
\toprule
Router & MiniLM & MPNet & Qwen & BGE & MBB & MBL & Mean & SD \\
\midrule
Zooter & 68.18 & 68.40 & 68.57 & 68.38 & 68.35 & 68.41 & 68.38 & 0.11 \\
kNN & 67.68 & 68.29 & 68.51 & 67.75 & 68.89 & 69.07 & 68.36 & 0.53 \\
Elo & 67.80 & 68.42 & 68.90 & 67.70 & 68.99 & 68.25 & 68.34 & 0.49 \\
Avengers-Pro & 68.27 & 68.20 & 68.50 & 68.22 & 68.68 & 67.99 & 68.31 & 0.22 \\
EquiRouter & 67.81 & 68.18 & 68.15 & 68.35 & 68.47 & 68.49 & 68.24 & 0.23 \\
KMeans & 68.26 & 68.58 & 68.30 & 68.01 & 68.88 & 68.25 & 68.38 & 0.28 \\
MESS+ & 68.09 & 68.17 & 68.08 & 68.12 & 68.30 & 68.44 & 68.20 & 0.13 \\
CARROT & 67.44 & 68.23 & 68.53 & 67.56 & 68.91 & 68.66 & 68.22 & 0.55 \\
CP-Router & 68.33 & 68.30 & 68.27 & 68.39 & 67.74 & 68.30 & 68.22 & 0.22 \\
P2L & 67.78 & 67.21 & 68.05 & 67.86 & 68.31 & 68.47 & 67.95 & 0.41 \\
MLP & 67.62 & 67.63 & 68.05 & 68.07 & 68.27 & 68.22 & 67.98 & 0.26 \\
EmbedLLM & 67.28 & 67.34 & 67.99 & 67.93 & 68.17 & 68.18 & 67.81 & 0.37 \\
ModelSpider & 67.56 & 67.44 & 67.76 & 67.75 & 67.63 & 67.50 & 67.61 & 0.12 \\
RoRF & 67.58 & 67.76 & 68.25 & 67.53 & --- & 68.18 & 67.86 & 0.30 \\
TagRouter & 67.34 & 67.34 & 68.69 & 67.34 & 68.83 & 68.30 & 67.97 & 0.65 \\
LinUCB & 67.58 & 68.20 & 67.78 & 67.77 & 67.74 & 67.51 & 67.76 & 0.22 \\
RouterDC & 67.35 & 67.59 & 67.69 & 67.40 & 67.53 & 67.67 & 67.54 & 0.13 \\
MIRT & 67.40 & 67.34 & 65.42 & 67.92 & 66.08 & 65.71 & 66.64 & 0.95 \\
IRT-Router & 67.36 & 67.34 & 64.17 & 67.34 & 67.35 & 65.35 & 66.49 & 1.27 \\
HybridLLM & 60.46 & 60.46 & 60.46 & 60.46 & 60.46 & 60.46 & 60.46 & 0.00 \\
OptLLM & --- & 66.55 & 66.63 & 66.52 & 66.86 & 66.48 & 66.61 & 0.14 \\
\bottomrule
\end{tabular}

\end{table}

\begin{table}[htbp]
\centering
\small
\setlength{\tabcolsep}{4pt}
\caption{\EmbedLLM{}: per-encoder test routing accuracy (\%). Mean and SD summarize the encoder columns; SD is in percentage points.}
\label{tab:app:encoders:embedllm}
\begin{tabular}{@{}lrrrrrrrr@{}}
\toprule
Router & MiniLM & MPNet & Qwen & BGE & MBB & MBL & Mean & SD \\
\midrule
Zooter & 66.86 & 67.35 & 67.75 & 67.43 & 67.80 & 67.86 & 67.51 & 0.35 \\
kNN & 66.30 & 67.90 & 67.80 & 65.63 & 68.00 & 67.30 & 67.16 & 0.89 \\
Elo & 66.97 & 66.87 & 67.90 & 65.87 & 67.90 & 66.60 & 67.02 & 0.72 \\
Avengers-Pro & 65.73 & 65.47 & 67.73 & 66.07 & 67.93 & 67.00 & 66.66 & 0.96 \\
EquiRouter & 66.36 & 66.73 & 67.07 & 67.13 & 67.29 & 67.87 & 67.08 & 0.47 \\
KMeans & 65.07 & 66.67 & 67.63 & 66.20 & 67.27 & 67.20 & 66.67 & 0.85 \\
MESS+ & 66.27 & 67.22 & 67.19 & 66.23 & 67.45 & 67.59 & 66.99 & 0.54 \\
CARROT & 66.57 & 67.27 & 67.53 & 64.73 & 68.23 & 67.57 & 66.98 & 1.12 \\
CP-Router & 65.87 & 66.77 & 65.20 & 65.93 & 65.63 & 67.17 & 66.09 & 0.67 \\
P2L & 64.59 & 65.95 & 66.63 & 65.89 & 67.51 & 67.69 & 66.38 & 1.06 \\
MLP & 65.43 & 65.97 & 66.69 & 66.43 & 67.18 & 67.38 & 66.51 & 0.67 \\
EmbedLLM & 64.29 & 65.37 & 66.52 & 66.06 & 66.85 & 67.19 & 66.05 & 0.98 \\
ModelSpider & 63.93 & 66.09 & 66.04 & 64.52 & 66.63 & 66.26 & 65.58 & 0.99 \\
RoRF & 64.43 & --- & --- & --- & --- & --- & 64.43 & 0.00 \\
TagRouter & 61.17 & 61.17 & --- & 60.90 & --- & --- & 61.08 & 0.13 \\
LinUCB & 64.07 & 65.80 & 66.83 & 65.57 & 65.80 & 66.50 & 65.76 & 0.88 \\
RouterDC & 57.93 & 58.21 & 58.68 & 57.14 & 58.32 & 58.69 & 58.16 & 0.53 \\
MIRT & 65.26 & 66.13 & 49.77 & 66.35 & 55.14 & 52.54 & 59.20 & 6.90 \\
IRT-Router & 58.53 & 60.51 & 46.41 & 59.18 & 43.96 & 43.75 & 52.06 & 7.42 \\
HybridLLM & 46.87 & 46.87 & 46.87 & 46.87 & 46.87 & 46.87 & 46.87 & 0.00 \\
OptLLM & --- & --- & --- & --- & --- & --- & --- & --- \\
\bottomrule
\end{tabular}

\end{table}

\begin{table}[htbp]
\centering
\small
\setlength{\tabcolsep}{4pt}
\caption{\RTwoBench{}: per-encoder test routing accuracy (\%). Mean and SD summarize the encoder columns; SD is in percentage points.}
\label{tab:app:encoders:r2bench}
\begin{tabular}{@{}lrrrrrrrr@{}}
\toprule
Router & MiniLM & MPNet & Qwen & BGE & MBB & MBL & Mean & SD \\
\midrule
Zooter & 81.99 & 82.09 & 82.21 & 82.11 & 82.21 & 82.09 & 82.12 & 0.08 \\
kNN & 81.97 & 82.01 & 82.37 & 81.90 & 82.16 & 82.08 & 82.08 & 0.15 \\
Elo & 82.01 & 82.08 & 82.37 & 81.86 & 82.05 & 82.13 & 82.08 & 0.15 \\
Avengers-Pro & 81.74 & 81.72 & 82.12 & 81.95 & 81.95 & 81.86 & 81.89 & 0.14 \\
EquiRouter & 81.26 & 81.72 & 82.08 & 81.87 & 81.88 & 81.99 & 81.80 & 0.26 \\
KMeans & 81.60 & 82.32 & 81.97 & 81.96 & 81.62 & 81.79 & 81.87 & 0.24 \\
MESS+ & 81.33 & 81.53 & 82.12 & 81.62 & 81.80 & 81.72 & 81.69 & 0.24 \\
CARROT & 81.40 & 81.81 & 81.57 & 81.28 & 81.45 & 81.73 & 81.54 & 0.19 \\
CP-Router & 81.72 & 82.13 & 82.25 & 82.13 & 81.61 & 81.76 & 81.93 & 0.24 \\
P2L & 80.98 & 81.39 & 82.09 & 81.62 & 81.65 & 81.61 & 81.56 & 0.33 \\
MLP & 80.87 & 81.20 & 81.88 & 81.00 & 81.72 & 81.62 & 81.38 & 0.38 \\
EmbedLLM & 80.46 & 81.32 & 81.92 & 81.58 & 81.45 & 81.69 & 81.40 & 0.46 \\
ModelSpider & 81.39 & 81.56 & 81.83 & 81.40 & 81.64 & 81.65 & 81.58 & 0.15 \\
RoRF & 81.36 & 81.64 & 82.23 & 81.14 & 81.51 & 81.73 & 81.60 & 0.34 \\
TagRouter & 81.10 & 81.10 & 82.06 & 81.10 & 81.67 & 81.40 & 81.41 & 0.36 \\
LinUCB & 80.36 & 80.97 & 81.07 & 80.82 & 80.60 & 80.47 & 80.72 & 0.26 \\
RouterDC & 79.75 & 79.71 & 80.27 & 79.81 & 81.27 & 80.51 & 80.22 & 0.55 \\
MIRT & 81.33 & 81.36 & 79.66 & 81.44 & 81.30 & 80.55 & 80.94 & 0.64 \\
IRT-Router & 81.21 & 81.35 & 80.01 & 81.14 & 81.33 & 81.20 & 81.04 & 0.47 \\
HybridLLM & 73.99 & 73.99 & 73.99 & 73.99 & 73.99 & 73.99 & 73.99 & 0.00 \\
OptLLM & 80.24 & 80.76 & 80.84 & 80.33 & 80.08 & 80.31 & 80.43 & 0.28 \\
\bottomrule
\end{tabular}

\end{table}

\begin{table}[htbp]
\centering
\small
\setlength{\tabcolsep}{4pt}
\caption{\CarrotB{}: per-encoder test routing accuracy (\%). Mean and SD summarize the encoder columns; SD is in percentage points.}
\label{tab:app:encoders:carrot}
\begin{tabular}{@{}lrrrrrrrr@{}}
\toprule
Router & MiniLM & MPNet & Qwen & BGE & MBB & MBL & Mean & SD \\
\midrule
Zooter & 86.51 & 86.69 & 86.65 & 86.61 & 86.70 & 86.67 & 86.64 & 0.06 \\
kNN & 86.59 & 87.13 & 86.82 & 86.65 & 86.54 & 86.97 & 86.78 & 0.21 \\
Elo & 86.59 & 87.18 & 87.04 & 86.68 & 86.78 & 86.86 & 86.85 & 0.20 \\
Avengers-Pro & 86.74 & 86.64 & 86.89 & 86.53 & 86.85 & 86.88 & 86.75 & 0.13 \\
EquiRouter & 86.10 & 86.48 & 86.57 & 86.45 & 86.53 & 86.63 & 86.46 & 0.17 \\
KMeans & 86.46 & 86.60 & 86.84 & 86.66 & 86.72 & 86.59 & 86.64 & 0.12 \\
MESS+ & 86.68 & 86.59 & 86.88 & 86.54 & 86.51 & 86.47 & 86.61 & 0.14 \\
CARROT & 86.08 & 86.47 & 86.44 & 86.40 & 86.02 & 86.83 & 86.37 & 0.27 \\
CP-Router & 86.77 & 86.74 & 87.02 & 86.81 & 86.29 & 86.27 & 86.65 & 0.28 \\
P2L & 85.89 & 86.17 & 86.62 & 86.19 & 86.52 & 86.53 & 86.32 & 0.26 \\
MLP & 85.87 & 85.88 & 86.32 & 85.92 & 86.22 & 86.49 & 86.12 & 0.24 \\
EmbedLLM & 85.85 & 86.42 & 86.33 & 86.70 & 86.34 & 86.39 & 86.34 & 0.25 \\
ModelSpider & 86.28 & 86.21 & 86.52 & 86.32 & 86.55 & 86.50 & 86.40 & 0.13 \\
RoRF & 85.68 & 85.97 & 86.75 & 86.30 & 86.44 & 86.40 & 86.26 & 0.34 \\
TagRouter & 84.82 & 84.82 & --- & --- & --- & --- & 84.82 & 0.00 \\
LinUCB & 85.55 & 85.87 & 86.19 & 86.33 & 86.04 & 85.22 & 85.87 & 0.38 \\
RouterDC & 83.06 & 82.97 & 83.20 & 83.05 & 83.23 & 83.29 & 83.13 & 0.11 \\
MIRT & 86.12 & 86.08 & 83.66 & 86.00 & 83.96 & 83.45 & 84.88 & 1.20 \\
IRT-Router & 85.52 & 85.78 & 83.69 & 86.04 & 83.97 & 83.68 & 84.78 & 1.02 \\
HybridLLM & 77.62 & 77.62 & 77.62 & 77.62 & 77.62 & 77.62 & 77.62 & 0.00 \\
OptLLM & 84.77 & 85.31 & 85.05 & 85.33 & 84.86 & 85.40 & 85.12 & 0.24 \\
\bottomrule
\end{tabular}

\end{table}

\FloatBarrier

\subsection{Benchmark Variations and Model-Pool Diversity}
\label{app:method-vs-benchmark}

We test whether the plateau persists under changes in query difficulty and task composition, and whether the candidate models offer complementary strengths.

\subsubsection{Test-set slicing}

We retain test queries that at least half of the candidate models answer correctly (task score above $0.5$), then rescore the available predictions from Tab.~\ref{tab:main:lambda0_all} with the same averaging and no retraining. Both router and oracle accuracies retain partial-credit scores. On these easier queries, the oracle gap shrinks, but the leading routers remain closely grouped and below the oracle (Tab.~\ref{tab:app:slices}).

\begin{table}[h]
\centering
\small
\caption{Full test set vs.\ queries that at least half of the pool solves. Gap is oracle minus best-router accuracy; spread is the top-five accuracy range on the filtered queries. Results use routers with available saved predictions, ranked separately on each subset.}
\label{tab:app:slices}
\setlength{\tabcolsep}{5pt}
\begin{tabular}{@{}lcccc@{}}
\toprule
 & Queries & \multicolumn{2}{c}{Gap to oracle (pp)} & Top-5 spread (pp) \\
\cmidrule(lr){3-4}
Benchmark & kept & full & filtered & filtered \\
\midrule
\RBench{} & 66.6\% & 11.30 & 2.82 & 0.03 \\
\NBshortk{} & 64.4\% & 14.27 & 5.67 & 0.23 \\
\EmbedLLM{} & 38.6\% & 30.46 & 4.64 & 1.44 \\
\RTwoBench{} & 67.5\% & 12.69 & 3.51 & 0.18 \\
\CarrotB{} & 77.5\% & 11.05 & 3.31 & 0.15 \\
\bottomrule
\end{tabular}%

\end{table}

\subsubsection{Task-domain groups}
\label{app:domain-plateau}

We group all 5,000 \NBshortk{} test queries by source-dataset domain and compare the same 12 available routers within each group, without retraining. In the four domains with at least 100 queries, the top-five spread is $0.22$--$1.22$~pp, while the best router remains $6.51$--$15.62$~pp below the oracle (Tab.~\ref{tab:app:domain-plateau}). Thus, the plateau also appears within task domains. The two smaller groups are reported for completeness.

\begin{table}[h]
\centering
\small
\caption{Domain breakdown on \NBshortk{} for the 12 methods with recoverable predictions from Tab.~\ref{tab:main:lambda0_all}. The oracle gap is measured from the best router within each domain; the top-five spread is the difference between the highest and fifth-highest router accuracies within that domain.}
\label{tab:app:domain-plateau}
\setlength{\tabcolsep}{5pt}
\begin{tabular}{@{}lrrr@{}}
\toprule
Task group & Queries & Oracle gap (pp) & Top-5 spread (pp) \\
\midrule
Math & 128 & 6.51 & 0.91 \\
Knowledge/commonsense/safety & 1,174 & 10.53 & 0.22 \\
Reading comprehension/QA & 3,387 & 15.62 & 0.22 \\
SuperGLUE & 278 & 14.81 & 1.22 \\
Code & 16 & 5.42 & 1.88 \\
Other specialized tasks & 17 & 6.67 & 2.55 \\
\midrule
All queries & 5,000 & 14.27 & 0.16 \\
\bottomrule
\end{tabular}%

\end{table}

\subsubsection{Retraining on filtered benchmarks}

We also filter each \NBshortk{} split to queries answered correctly by at least three or five models, using the same correctness threshold as above. We retrain the six routers in Tab.~\ref{tab:app:rebuilt} with ModernBERT-base and MiniLM-L6 over three seeds, comparing them with full-data training on the same filtered test queries. Filtering the training data yields little improvement for five routers and lowers MLP accuracy; the best router still trails the oracle by $5.5$--$9.7$~pp.

\begin{table}[h]
\centering
\small
\caption{Test accuracy (\%) after training on filtered \NBshortk{} data, averaged over two encoders and three seeds. $\Delta$ is the change from full-data training, in pp, evaluated on the same filtered queries. Original task scores are retained, so passing the correctness threshold does not imply an oracle score of $100\%$.}
\label{tab:app:rebuilt}
\setlength{\tabcolsep}{5pt}
\begin{tabular}{@{}lcccc@{}}
\toprule
 & \multicolumn{2}{c}{$\ge3$ models correct ($3{,}663$ queries)} & \multicolumn{2}{c}{$\ge5$ models correct ($3{,}219$ queries)} \\
\cmidrule(lr){2-3} \cmidrule(lr){4-5}
Router & Accuracy & $\Delta$ & Accuracy & $\Delta$ \\
\midrule
KMeans & 89.53 & $+0.03$ & 93.81 & $+0.10$ \\
Elo & 89.38 & $-0.06$ & 93.42 & $-0.32$ \\
kNN & 89.33 & $+0.11$ & 93.38 & $-0.20$ \\
RouterDC & 88.06 & $-0.17$ & 92.47 & $-0.19$ \\
MIRT & 87.28 & $+0.18$ & 92.21 & $+0.08$ \\
MLP (CE) & 85.11 & $-2.20$ & 91.06 & $-1.12$ \\
\midrule
Oracle   & \multicolumn{2}{c}{99.22} & \multicolumn{2}{c}{99.35} \\
\bottomrule
\end{tabular}%

\end{table}

\subsubsection{Model-pool diversity and specialization}
\label{app:pool-diversity}

The oracle exceeds the best single model by $11.4$--$37.2$~pp across the five benchmarks. Thus, the plateau occurs in pools where selecting different models for different queries offers substantial potential gains.

Together, these checks support the main-text finding that leading routers remain closely grouped below the oracle across the benchmark variations we test, even when the candidate models have complementary strengths.

\subsection{Multi-Sample Correctness Labels}
\label{app:multisample}

To assess the sensitivity of the oracle to single-sample response labels, we collected ten independent responses for each of the nine models on the same 5{,}000 \NBshortk{} test queries and aggregated their correctness by majority vote. API requests used provider defaults without temperature or top-$p$ overrides, matching the original collection.

Tab.~\ref{tab:app:multisample} reports the resulting majority-label oracles. Its gap column subtracts the original best-router accuracy, $0.684$, as a fixed reference. The majority-label oracle remains close to the single-sample oracle, supporting the observation that much of the pool's apparent routing opportunity survives repeated sampling.

\begin{table}[h]
\centering
\small
\caption{Oracle accuracy on \NBshortk{} under single-sample and majority-vote correctness, with the original best-router accuracy ($0.684$, Tab.~\ref{tab:main:lambda0_all}) held fixed as a reference.}
\label{tab:app:multisample}
\setlength{\tabcolsep}{8pt}
\begin{tabular}{@{}lcc@{}}
\toprule
Correctness label & Oracle & Gap to fixed reference \\
\midrule
Single sample (labels used in this paper) & 0.827           & 14.3 pp \\
$k{=}10$ majority                   & 0.817    & 13.3 pp \\
\bottomrule
\end{tabular}%

\end{table}

\FloatBarrier

\section{Analysis Details}

\subsection{Hard-Query Analysis and Model-Selection Distributions}
\label{app:more-examples}
The behavioral analyses use the 12 routers shown in Fig.~\ref{fig:score_spread_comparison} and below, with ModernBERT-base, seed 42, and validation-selected architectures. We convert task scores to binary correctness labels, $y_{im}=\mathbf{1}[s_{im}\ge0.5]$, and use the hard/easy split from Sec.~\ref{sec:plateau-stratification}. In Tab.~\ref{tab:strata}, the hard-query gap share is the hard-query fraction multiplied by the hard-subset gap, divided by the full-test gap.

The figures extend the main-text comparison to nine additional routers on all five benchmarks. Each router histogram shows its selection frequencies; ``Truth'' shows each model's share of all correct query--model pairs on hard queries. Models are ordered by training accuracy, with the dotted line marking the top-ranked model. Router selections remain concentrated on globally strong models, while correct answers on hard queries are more evenly spread across the pool.

Rows show (Elo, P2L, Zooter), (CARROT, CP-Router, Avengers-Pro), and (EmbedLLM, KMeans, EquiRouter), using the same panel format as Fig.~\ref{fig:score_spread_comparison}.

\begin{figure}[h]
\centering
\includegraphics[width=\textwidth]{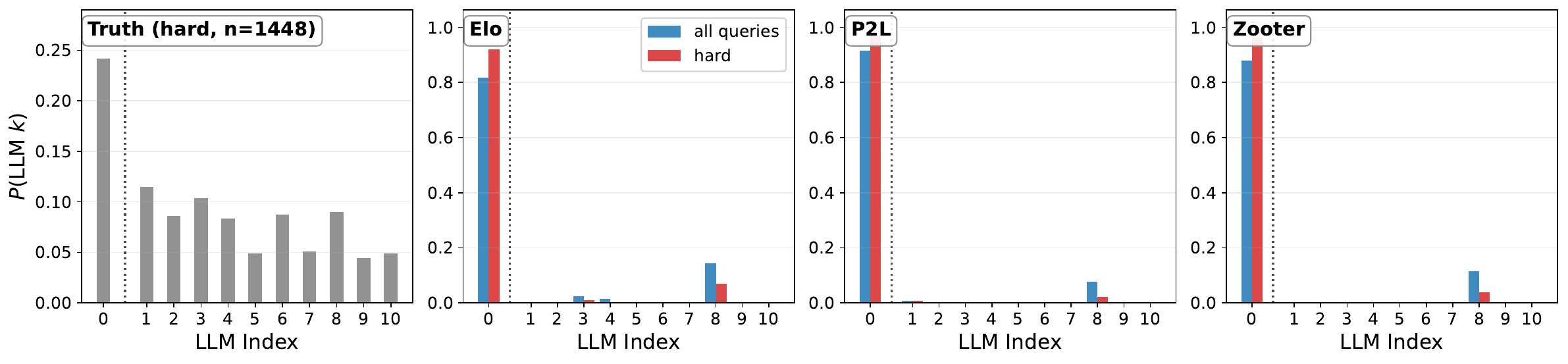}\\[2pt]
\includegraphics[width=\textwidth]{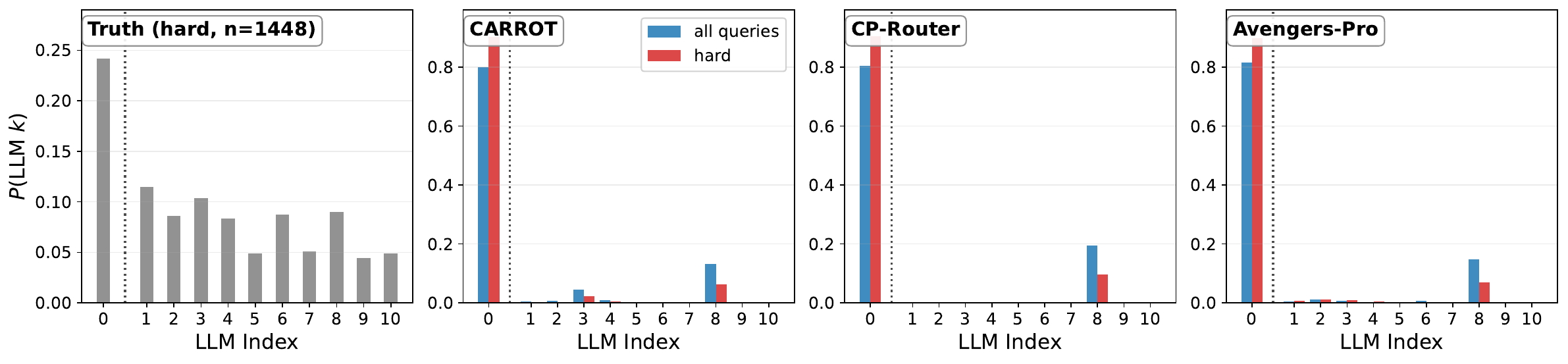}\\[2pt]
\includegraphics[width=\textwidth]{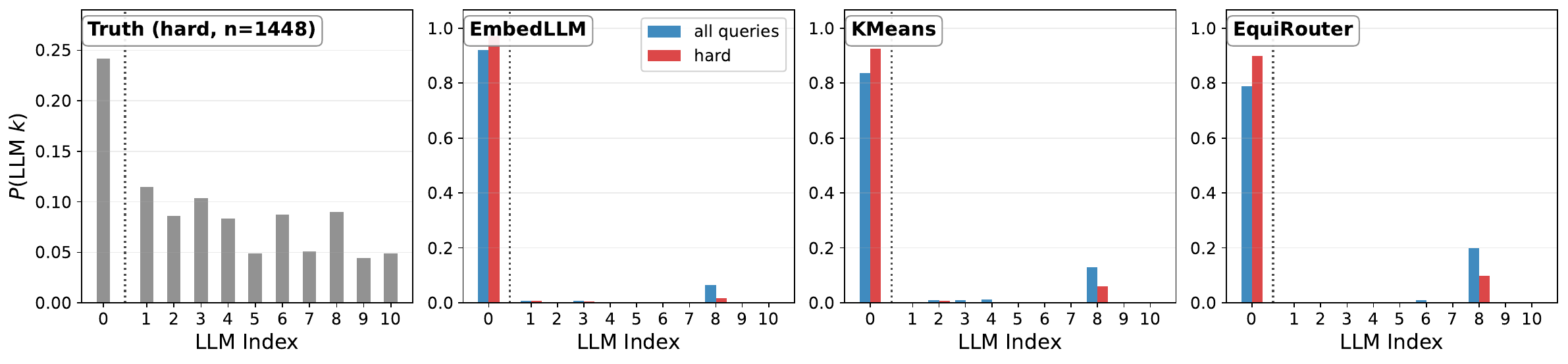}
\caption{\RBench{}: selection frequency on hard vs.\ all queries for nine additional routers.}
\label{fig:app:rb_extra}
\end{figure}

\begin{figure}[h]
\centering
\includegraphics[width=\textwidth]{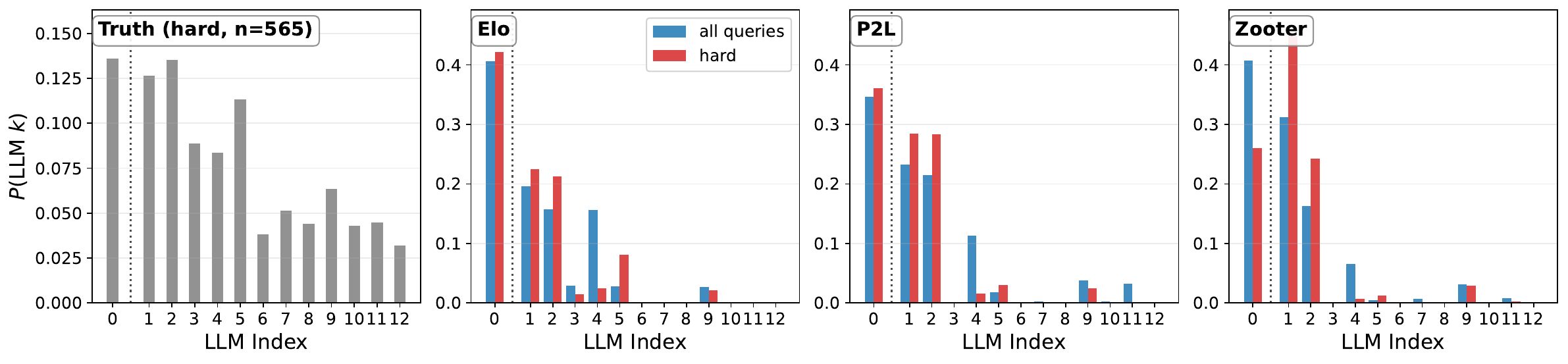}\\[2pt]
\includegraphics[width=\textwidth]{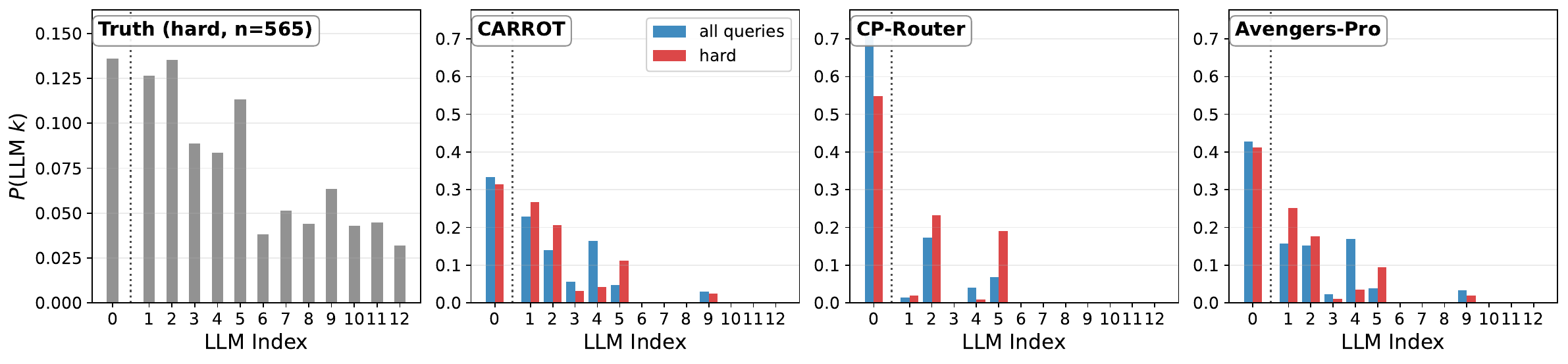}\\[2pt]
\includegraphics[width=\textwidth]{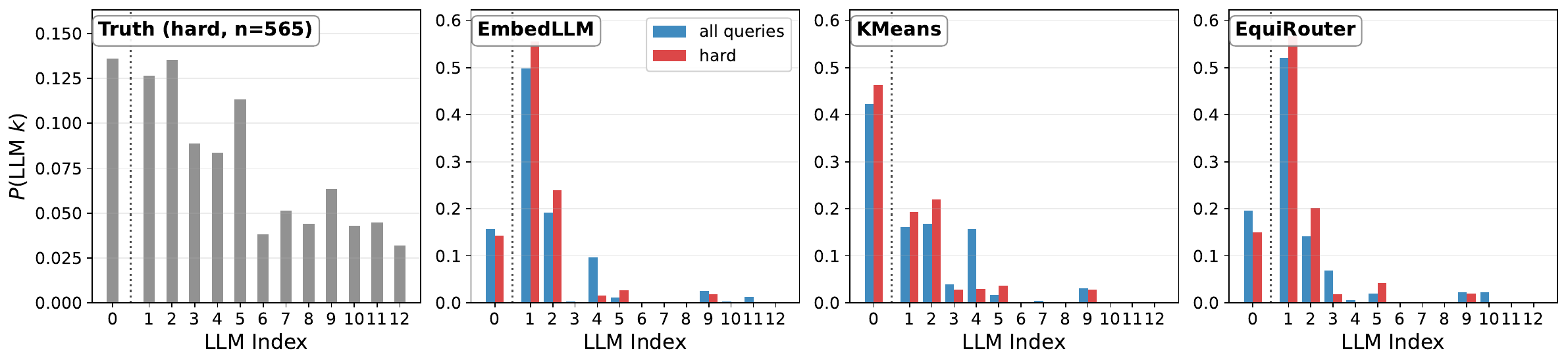}
\caption{\CarrotB{}: selection frequency on hard vs.\ all queries for nine additional routers.}
\label{fig:app:carrot_extra}
\end{figure}

\begin{figure}[h]
\centering
\includegraphics[width=\textwidth]{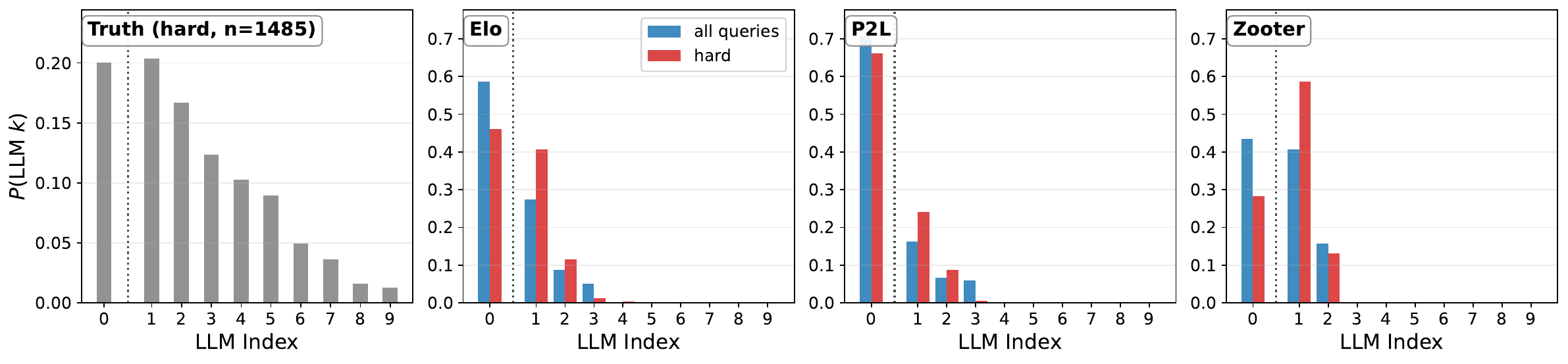}\\[2pt]
\includegraphics[width=\textwidth]{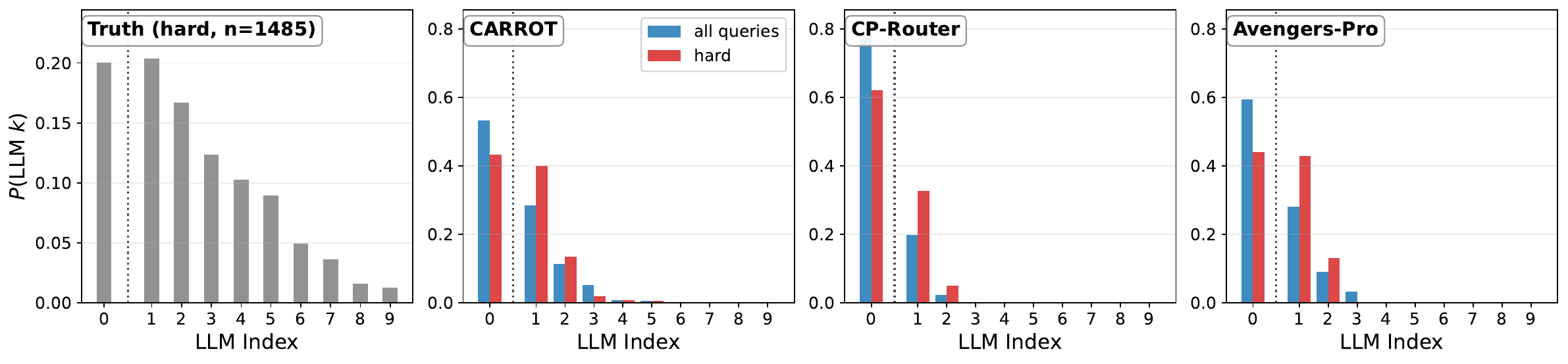}\\[2pt]
\includegraphics[width=\textwidth]{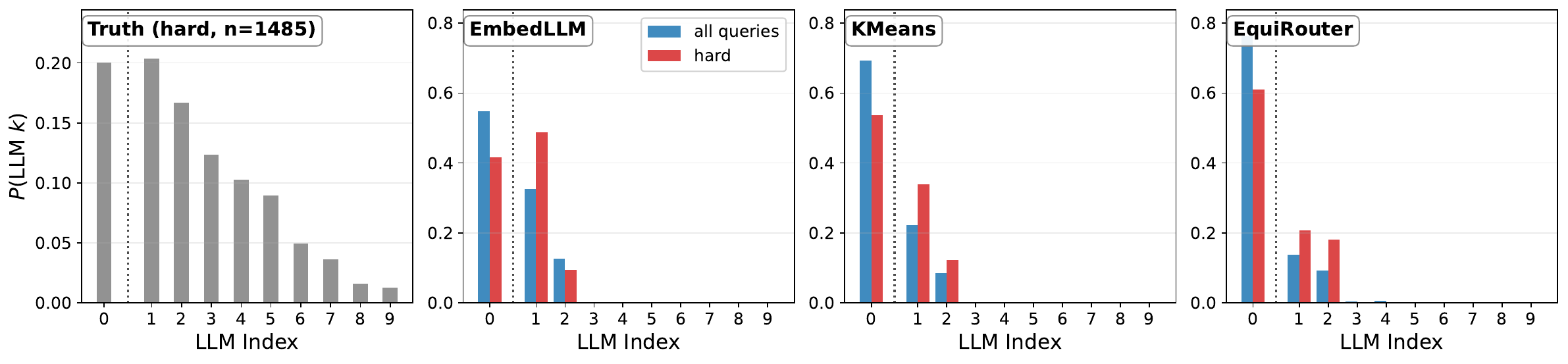}
\caption{\RTwoBench{}: selection frequency on hard vs.\ all queries for nine additional routers.}
\label{fig:app:r2bench_extra}
\end{figure}

\begin{figure}[h]
\centering
\includegraphics[width=\textwidth]{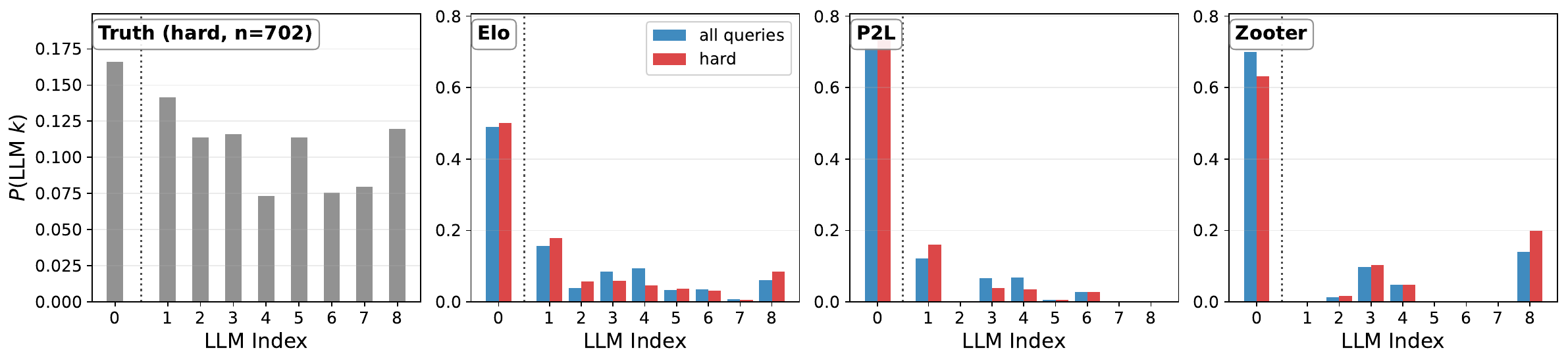}\\[2pt]
\includegraphics[width=\textwidth]{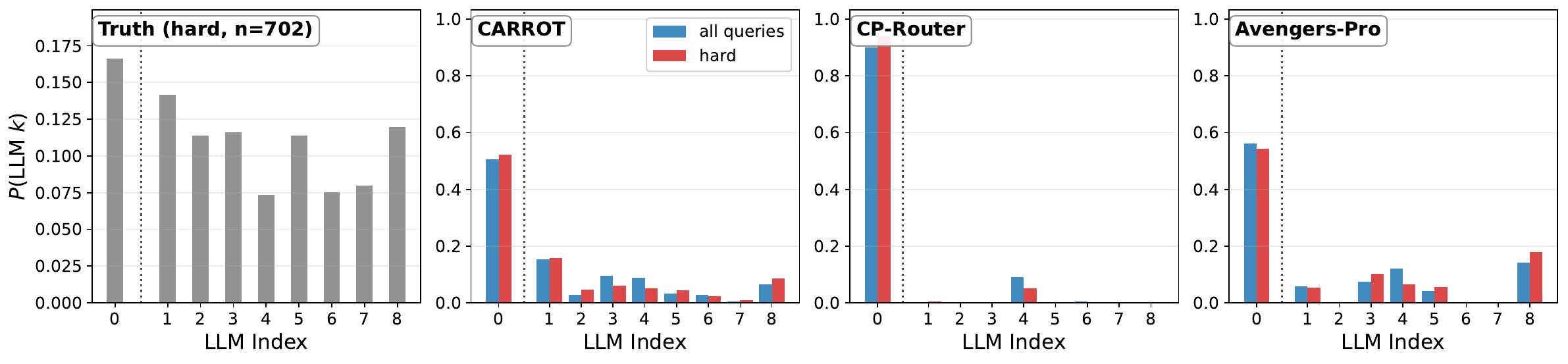}\\[2pt]
\includegraphics[width=\textwidth]{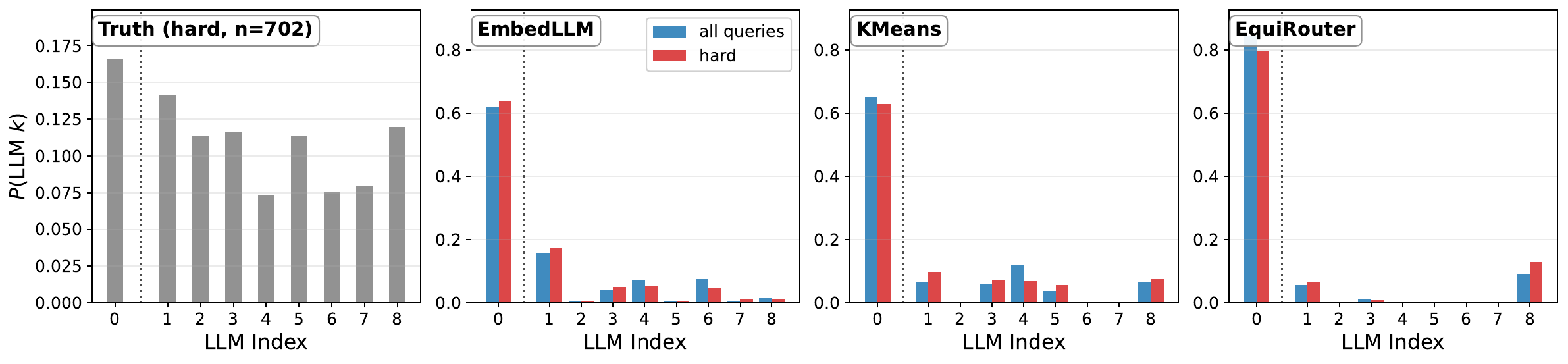}
\caption{\NBshortk{}: selection frequency on hard vs.\ all queries for nine additional routers.}
\label{fig:app:nb30k_extra}
\end{figure}

\begin{figure}[h]
\centering
\includegraphics[width=\textwidth]{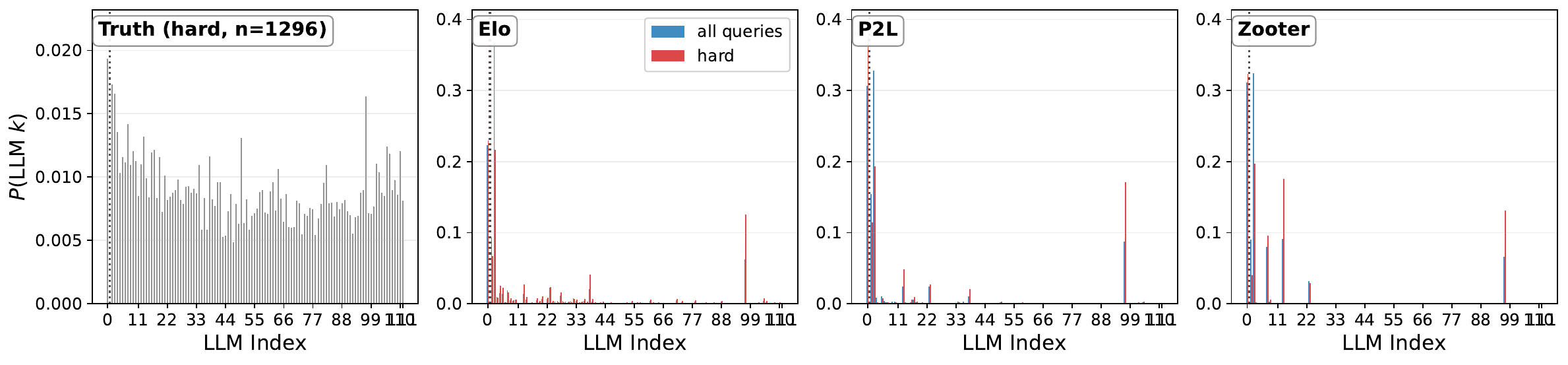}\\[2pt]
\includegraphics[width=\textwidth]{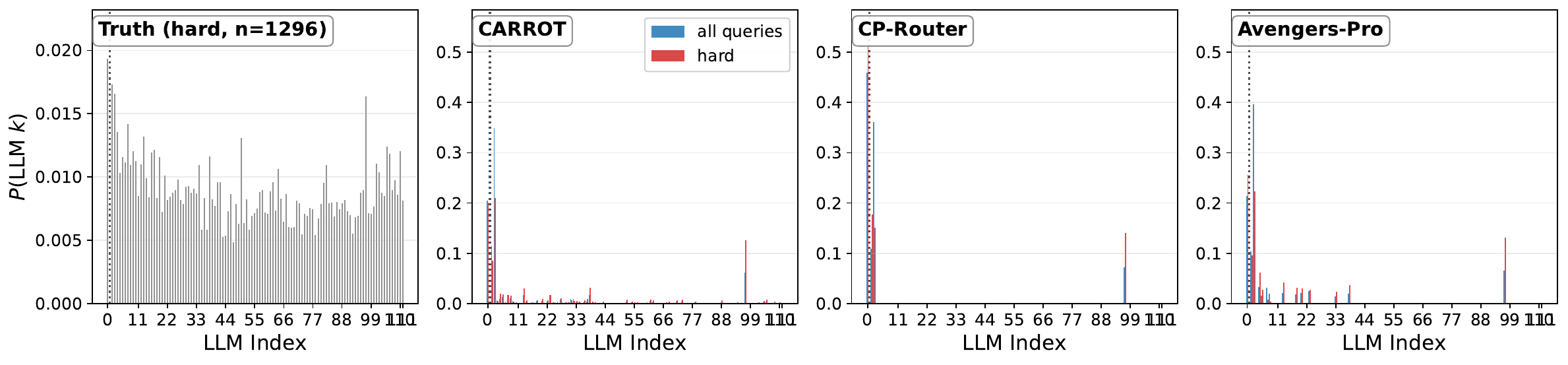}\\[2pt]
\includegraphics[width=\textwidth]{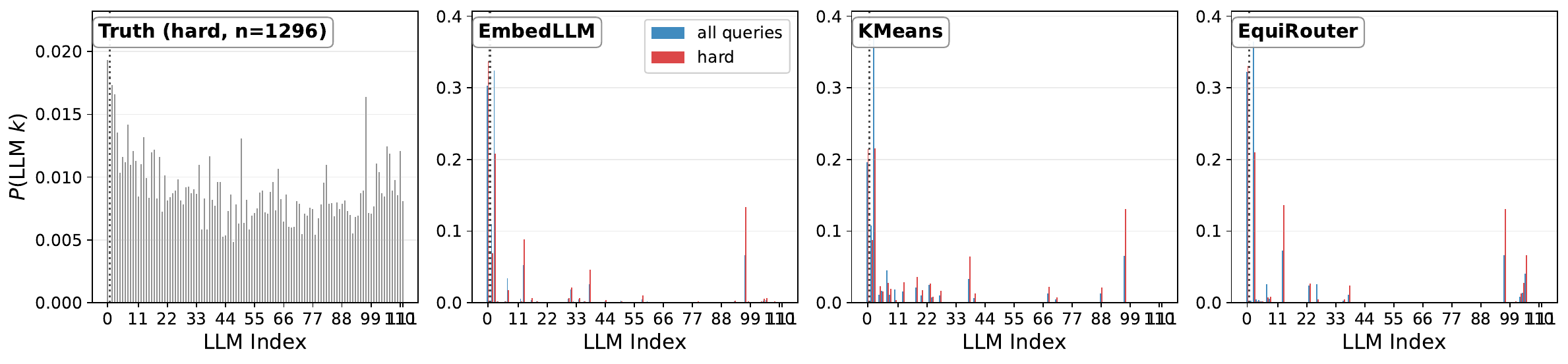}
\caption{\EmbedLLM{}: selection frequency on hard vs.\ all queries for nine additional routers.}
\label{fig:app:embedllm_extra}
\end{figure}

\FloatBarrier
\clearpage

\subsection{Pairwise Error Exchange}
\label{app:exchange-details}
We compare all 66 pairs of the 12 routers in App.~\ref{app:more-examples}. For each pair, we identify queries where the routers select different models. Within these disagreements, unsigned exchange is the fraction on which exactly one router succeeds; signed exchange is the difference between their win rates. Tab.~\ref{tab:exchange} reports the range of disagreement rates and medians across pairs for the other metrics. Each exchange ratio is computed before taking the median, so it need not equal the ratio of the reported medians.

\subsection{kNN Comparison Setup}
\label{app:knn-comparison}

We compute concordance as defined in Sec.~\ref{sec:prelim}, using the diagnostic runs in App.~\ref{app:more-examples}. The comparison includes kNN and 11 other routers. The hard-query median uses nine routers: MLP, Elo, MESS+, P2L, Zooter, CARROT, CP-Router, Avengers-Pro, and EmbedLLM.

\FloatBarrier

\section{Scaling and Fine-Tuning}

\subsection{Scaling and Fine-Tuning Setup}
\label{app:scaling-setup}
The eight Scaled-FT runs jointly train the encoder and routing head with seed 42, batch size 32, encoder/head learning rates of $2\times10^{-5}$/$10^{-3}$, weight decay of $0.01$, and gradient clipping at norm 1.0. The frozen kNN reference uses the scaled training set and ModernBERT-large embeddings with $k=750$.

\subsection{Encoder Scaling and Fine-Tuning on Other Benchmarks}
\label{app:fix-crossbench}

On \CarrotB{} and \EmbedLLM{}, we keep the original training sets and compare ModernBERT-base (mbb) and ModernBERT-large (mbl), each frozen or fine-tuned, using the same eight routers as Sec.~\ref{sec:fix}.

\begin{table}[!htbp]
\centering
\footnotesize
\caption{Encoder scaling and fine-tuning on \CarrotB{} and \EmbedLLM{}. Means are across eight routers; best FT and best kNN are the strongest configurations across encoder choices. $\Delta$ is best FT minus best kNN, in pp.}
\label{tab:app:fix-crossbench}
\setlength{\tabcolsep}{2.4pt}
\begin{tabular}{@{}lccccccc@{}}
\toprule
Benchmark   & frzn mbb & frzn mbl & FT mbb & FT mbl & best FT cell                       & best kNN ref      & $\Delta$ pp \\
\midrule
\CarrotB{}  & 0.8604   & 0.8609   & 0.8645 & 0.8654 & $0.8711$ (\texttt{routerdc}, mbl)  & $0.8697$ (mbl)    & $+0.14$     \\
\EmbedLLM{} & 0.6600   & 0.6643   & 0.6779 & 0.6764 & $0.6817$ (\texttt{routerdc}, mbb)  & $0.6800$ (mbb)    & $+0.17$     \\
\bottomrule
\end{tabular}
\end{table}

Fine-tuning the large encoder improves mean accuracy over the frozen-base baseline by $0.50$ and $1.64$~pp, respectively. The best fine-tuned routers remain close to frozen kNN, exceeding it by only $0.14$ and $0.17$~pp.

\end{document}